\documentclass{article}

\usepackage{arxiv}

\usepackage[utf8]{inputenc} %
\usepackage[T1]{fontenc} %
\usepackage{hyperref} %

\usepackage{amsmath}
\usepackage{amssymb}
\usepackage{amsfonts}
\usepackage[english]{babel}
\usepackage{verbatim}
\usepackage{textcomp}

\usepackage{tabularx}

\usepackage{acronym}

\usepackage{booktabs}

\usepackage{placeins}

\usepackage{pgfplots}
\usepgflibrary{fpu} 
\pgfplotsset{minor grid style=dotted, table/search path={plots/}}

\pgfplotsset{select coords between index/.style 2 args={
		x filter/.code={
			\ifnum\coordindex<#1\fi
			\ifnum\coordindex>#2\fi
		}
}}

\usepgfplotslibrary{fillbetween}

\graphicspath{{images/}}
\usepackage{subcaption}
\usepackage[export]{adjustbox}
\usepackage{tikz}
\usetikzlibrary{external}
\usepackage{tikz-3dplot}
\usetikzlibrary{positioning,shapes,arrows,calc,3d,intersections,decorations.pathreplacing,angles,quotes,shapes.misc,decorations.markings,chains}
\usetikzlibrary{matrix}
\usetikzlibrary{arrows.meta}

\tikzset{%
	>={Latex[width=2mm,length=2mm]},
	base/.style = {rectangle, rounded corners, draw=black,
		minimum width=2.5cm, minimum height=0.5cm,
		text centered, font={\sffamily,\scriptsize}},
	activityStarts/.style = {base, fill=stevens2!30},
	startstop/.style = {base, fill=red!30},
	activityRuns/.style = {base, fill=green!30},
	process/.style = {base, minimum width=3.5cm, fill=stevens1!30,},
	failureMode/.style = {text centered, font={\sffamily,\scriptsize}},
	listing/.style = {base, minimum width=2.0cm,font={\sffamily,\tiny}},
	outer/.style = {base, double distance =0.5pt, minimum width=2.5cm,font={\sffamily,\tiny}},
}

\usepackage{xcolor}

\definecolor{stevens1}{RGB}{163,38,56}
\definecolor{stevens2}{RGB}{0,67,128}
\definecolor{stevens3}{RGB}{249,142,43}
\definecolor{stevens4}{RGB}{3,87,19}
\definecolor{stevens_gray}{RGB}{127,127,127}
\definecolor{stevens_medium_blue}{RGB}{72,150,207}
\definecolor{stevens_medium_orange}{RGB}{231,132,46}
\definecolor{stevens_light_gray}{RGB}{228,229,230}

\tikzset{%
	>={Latex[width=2mm,length=2mm]},
	base/.style = {rectangle, rounded corners, draw=black,
		minimum width=2.5cm, minimum height=0.5cm,
		text centered, font={\sffamily,\scriptsize}},
	activityStarts/.style = {base, fill=stevens2!30},
	startstop/.style = {base, fill=red!30},
	activityRuns/.style = {base, fill=green!30},
	process/.style = {base, minimum width=3.5cm, fill=stevens1!30,},
	failureMode/.style = {text centered, font={\sffamily,\scriptsize}},
	listing/.style = {base, minimum width=2.0cm,font={\sffamily,\tiny}},
	outer/.style = {base, double distance =0.5pt, minimum width=2.5cm,font={\sffamily,\tiny}},
}

\usepackage{bm}

\usepackage[stable]{footmisc}

\usepackage[acronym,shortcuts,nomain]{glossaries}
\makeglossaries

\newacronym{XFEM}{XFEM}{Extended Finite Element Method}
\newacronym{CDM}{CDM}{Continuum Damage Mechanics}
\newacronym{UEL}{UEL}{User Element Subroutine}
\newacronym{CZM}{CZM}{Cohesive Zone Model}
\newacronym{USDFLD}{USDFLD}{User subroutine to define field variables at a material point}
\newacronym{UDMGINI}{UDMGINI}{User subroutine to define damage initiation criterion}
\newacronym{SDVINI}{SDVINI}{User subroutine to define initial solution-dependent state variable fields}

\newacronym{UC}{UC}{Uncertainty Quantification}

\newacronym{SFEM}{SFEM}{Stochastic Finite Element Method}
\newacronym{RFEM}{RFEM}{Random Finite Element Method}
\newacronym{PDE}{PDE}{Partial Differential Equation}
\newacronym{SSFEM}{SSFEM}{Spectral Stochastic Finite Element Method}
\newacronym{XSFEM}{XSFEM}{Extended Stochastic Finite Element Method}

\newacronym{QMC}{QMC}{Quasi Monte Carlo}
\newacronym{SEM}{SEM}{Scanning Electron Microscope}
\newacronym{FORM}{FORM}{First-Order Reliability Method}
\newacronym{SORM}{SORM}{Second-Order Reliability Method}

\newacronym{LDS}{LDS}{Low Discrepancy Sequence}
\newacronym{LHS}{LHS}{Latin Hypercube Sampling}

\newacronym{FEM}{FEM}{Finite Element Method}
\newacronym{DIC}{DIC}{Digital Image Correlation}
\newacronym{CNN}{CNN}{Convolutional Neural Network}
\newacronym{KLE}{KLE}{Karhunen-Lo\`{e}ve Expansion}
\newacronym{NNW}{NNW}{Neural Network}
\newacronym{ReLu}{ReLu}{Rectified Linear Unit}
\newacronym{FDM}{FDM}{Fused Deposition Modeling}
\newacronym{MC}{MC}{Monte Carlo}
\newacronym{ENAS}{ENAS}{Efficient Neural Architecture Search}
\newacronym{RNN}{RNN}{Recurrent Neural Network}
\newacronym{NAS}{NAS}{Neural Architecture Search}
\newacronym{DAG}{DAG}{Directed Acyclic Graph}
\newacronym{LSTM}{LSTM}{Long Short-Term Memory}
\newacronym{GRU}{GRU}{Gated Recurrent Unit}
\newacronym{GRN}{GRN}{Gated Residual Network}
\newacronym{GLU}{GLU}{Gated Linear Unit}
\newacronym{ELU}{ELU}{Exponential Linear Unit}
\newacronym{VDIC}{VDIC}{Virtual Digital Image Correlation}
\newacronym{UQ}{UQ}{Uncertainty Quantification}
\newacronym{AM}{AM}{Additively Manufactured}
\newacronym{DNN}{DNN}{Deep Neural Network}
\newacronym{AI}{AI}{Artificial Intelligence}
\newacronym{NNI}{NNI}{Neural Network Intelligence}
\newacronym{PCE}{PCE}{Polynomial Chaos Expansion}

\newacronym{FAA}{FAA}{Federal Aviation Administration}

\newacronym{ROM}{ROM}{Reduced Order Model}
\newacronym{MFH}{MFH}{Mean-Field Homogenization}
\newacronym{RVE}{RVE}{Representative Volume Element}
\newacronym{SVE}{SVE}{Stochastic Volume Element}
\newacronym{BI}{BI}{Bayesian Interference}
\newacronym{SFRP}{SFRP}{Short-Fiber Reinforced Composite}
\newacronym{ML}{ML}{Machine Learning}
\newacronym{BNN}{BNN}{Bayesian Neural Network}
\newacronym{GPR}{GPR}{Gaussian Process Regression}
\newacronym{MD}{MD}{Moldecular Dynamics}
\newacronym{micro-CT}{micro-CT}{Microcomputed Tomography}
\newacronym{RL}{RL}{Reinforcement Learning}

\newacronym{DARTS}{DARTS}{Differentiable Architecture Search}
\newacronym{FFF}{FFF}{Fused Filament Fabrication}
\newacronym{MSE}{MSE}{Mean Squared Error}
\newacronym{SPOS}{SPOS}{Single Path One-Shot Neural Architecture Search}

\newacronym{NN}{NN}{Neural Network}
\newacronym{GNN}{GNN}{Graph Neural Network}
\newacronym{DUNN}{DUNN}{Dropout Neural Network}
\newacronym{DMN}{DMN}{Deep Material Network}
\newacronym{SNR}{SNR}{Signal-to-Noise Ratio}
\newacronym{UP}{UP}{Uncertainty Propagation}
\newacronym{PDF}{PDF}{Probability Distribution Function}

\newacronym{NLP}{NLP}{Natural Language Processing}
\newacronym{GPT}{GPT}{Generative Pre-trained Transformer}
\newacronym{PCA}{PCA}{Principal Component Analysis}
\newacronym{PINN}{PINN}{Physics-Informed Neural Network}
\newacronym{POD}{POD}{Proper Orthogonal Decomposition}
\newacronym{FVR}{FVR}{Fiber Volume Ratio}
\newacronym{PBC}{PBC}{Periodic Boundary Condition}

\newacronym{BVP}{BVP}{Boundary Value Problem}

\newacronym{PyMKS}{PyMKS}{Materials Knowledge Systems in Python}
\newacronym{LLM}{LLM}{Large Language Model}
\newacronym{RMSE}{RMSE}{Root-Mean-Square Error}

\title{A Neural Network Transformer Model for Composite Microstructure Homogenization}

\author{ \href{https://orcid.org/0000-0001-8860-3014}{\includegraphics[scale=0.06]{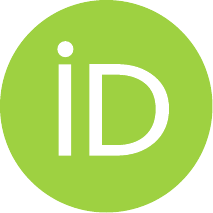}\hspace{1mm}Emil Pitz}\\
	Department of Mechanical Engineering\\
	Stevens Institute of Technology\\
	Hoboken, NJ 07030, USA\\
	\texttt{emil.pitz@gmail.com} \\
	\And
	\href{https://orcid.org/0000-0002-0248-8658}{\includegraphics[scale=0.06]{orcid.pdf}\hspace{1mm}Kishore Pochiraju}\\
	Department of Mechanical Engineering\\
	Stevens Institute of Technology\\
	Hoboken, NJ 07030, USA\\
	\texttt{kpochira@stevens.edu} \\
}

\renewcommand{\headeright}{AM}
\renewcommand{\undertitle}{Accepted Manuscript}
\renewcommand{\shorttitle}{A Transformer Model for Composite Microstructure Homogenization}

\hypersetup{
	pdftitle={A Neural Network Transformer Model for Composite Microstructure Homogenization},
	pdfauthor={Emil Pitz, Kishore Pochiraju},
	pdfkeywords={Transformer, Microstructure Homogenization, Surrogate Model, History-dependence, Microstructure encoding},
}

\begin{document}
	\maketitle

		\begin{abstract}
			Heterogeneity and uncertainty in a composite microstructure lead to either computational bottlenecks if modeled rigorously or to solution inaccuracies in the stress field and failure predictions if approximated. Although methods suitable for analyzing arbitrary and non-linear microstructures exist, their computational cost makes them impractical to use in large-scale structural analysis. Surrogate models or \acp*{ROM} commonly enhance efficiencies but are typically calibrated with a single microstructure. Homogenization methods, such as the Mori-Tanaka method, offer rapid homogenization for a wide range of constituent properties. However, simplifying assumptions, like stress and strain averaging in phases, render the consideration of both deterministic and stochastic variations in microstructure infeasible.
			This paper illustrates a transformer neural network architecture that captures the knowledge of various microstructures and constituents, enabling it to function as a computationally efficient homogenization surrogate model. Given an image or an abstraction of an arbitrary composite microstructure of linearly elastic fibers in an elastoplastic matrix, the transformer network predicts the history-dependent, non-linear, and homogenized stress-strain response. Two methods for encoding microstructure features were tested: calculating two-point statistics using \ac*{PCA} for dimensionality reduction and employing an autoencoder with a \ac*{CNN}. Both methods accurately predict the homogenized material response.
			The developed transformer neural network offers an efficient means for microstructure-to-property translation, generalizable and extendable to a variety of microstructures. The paper describes the network architecture, training and testing data generation, and performance under cycling and random loadings.

			\end{abstract}
		

	\acresetall
	\section{Introduction}
	
Multiple studies have shown that \acp*{NN} can replace constitutive laws of homogeneous materials for various material behaviors, such as plastic behavior \cite{zhang_using_2020,bessa_framework_2017}, plastic material under cyclic loading \cite{furukawa_accurate_2004}, hyperelastic materials \cite{chen_polyconvex_2022}, or for modeling traction-separation behavior \cite{wang_meta-modeling_2019}. Extensions of \ac*{NN} models that predict the orthotropic linear elastic properties of 2D and 3D-microcells have been shown in \cite{messner_convolutional_2020,rao_three-dimensional_2020,peng_ph-net_2022} using \acp*{CNN}. A surrogate model for the anisotropic electrical response of graphene/polymer nanocomposites was presented in \cite{lu_data-driven_2019}.

To model the non-linear response of heterogeneous materials, \ac*{NN} surrogates of a non-linear strain energy function were presented in \cite{le_computational_2015,nguyen-thanh_surrogate_2019}. In \cite{mianroodi_teaching_2021,khorrami_artificial_2022}, localized Mises stress distributions were generated using \acp*{CNN} for a given monotonic uniaxial tensile loading at a prescribed loading rate. The approaches in \cite{mianroodi_teaching_2021,khorrami_artificial_2022} were limited to monotonic loading and history dependence was ignored. The loading history dependence can be significant for the homogenized material level to find the stress state for irreversible behavior under general loading conditions \cite{wu20}. Therefore, in \cite{settgast_hybrid_2020} state variables were used to account for the loading history and build a surrogate for the yield function of foams. An alternative to using state variables is a \acp*{NN} that takes into account of entire loading history. An \ac*{LSTM} based \ac*{RNN} that considers strain history and microstructure information of elastoplastic composites was described in \cite{mozaffar19}. Similarly, a \ac*{RNN} surrogate was implemented in \cite{ghavamian_accelerating_2019} for history-dependent homogenization to accelerate nested (micro-macro) finite element (FE$^2$) analysis with automatic differentiation of the surrogate model to obtain consistent tangent matrices. In \cite{wu20} a \ac*{RNN} with \acp*{GRU} was used to implement a surrogate model for elastoplastic composite microcells. The methodology was extended to include localization in the surrogate model to extract the micro stress field using dimensionality reduction with \ac*{PCA} \cite{wu_recurrent_2022}.

In this paper, we described a surrogate model that can predict the response of arbitrary microstructures with two constituent phases. The model can predict the current (history-dependent, e.g., due to plasticity) stress state average over the \ac*{RVE} $\bm{S}_M$ at time $t$ as a result of the strain history $\bm{E}_{M,1:t}$ for a given microstructure defined by parameters $\bm{m}$ and constituent properties $\bm{p}$ \cite{mozaffar19}:

 \begin{equation}
	\bm{S}_M = f\left( \bm{E}_{M,1:t}, \bm{m}, \bm{p}, t\right).
\end{equation}

 The transformer network architecture was developed in \cite{vasvani17} was used as an alternative to \acp*{RNN}. Transformer NNs have been shown to have impressive performance in language modeling tasks, machine translation and a variety of other applications (see Sec.~\ref{sec:transformer_intro}). To date, transformers have not yet been used as surrogate models for a history-dependent material response. 

Transformers in the context of \ac*{NLP} applications when trained on diverse data have shown an impressive ability to generalize. A single model has been able to generate text, and write and debug code in different programming languages \cite{radford_language_2019}. We believe that the ability of transformers to generalize can be extended to the field of microstructure homogenization. We hypothesized that a transformer could be trained on a variety of material combinations and microstructures and function as a fast and efficient homogenization tool. Because of the prediction speed, the trained model could be used to perform \ac*{UQ} of the response of composite structures with complex and varying microstructure, where current homogenization techniques still are a computational bottleneck. 

In this effort, the feasibility of using a transformer network to predict the history-dependent material response was first trained and validated with homogeneous materials. Following the successful prediction of the history-dependent homogeneous material response, the surrogate model was extended and trained for heterogeneous materials.
Furthermore, in \ac*{NLP}, pre-training of transformers, where the network is first trained on large unlabeled datasets followed by task-specific fine-tuning, has shown major performance advantages \cite{devlin_bert_2019}. Following a similar strategy, we used the weights for homogeneous data as a starting point for training with heterogeneous materials. This strategy reduced the amount of computation and the microstructure homogenization data required for training the transformer network.

The structure of the remainder of the paper is as follows. First, we will provide a literature review and introduce transformer neural network architectures to learn sequence-to-sequence relationships in Section~\ref{sec:transformer_intro}. The application to the micro-structure homogenization problem and generation of training data for a non-linear, history-dependent, and two-phase composite is described in Section~\ref{sec:training_data}. Introducing dimensionality reduction methods is the focus of the \ref{sec:microstructure_representation}. Finally, the transformer architecture with embedded microstructure information and the training procedure are explained in Section~\ref{sec:surrogate}, followed by a discussion of the surrogate model performance.

\section{Transformer Neural Network Architectures for Sequence-to-Sequence Learning}\label{sec:transformer_intro}
Transformer models are recent compared to \acp*{RNN} to analyze sequential data. First presented in \cite{vasvani17}, transformers have been successful in \ac*{NLP} applications and, more recently, have been applied to a wider range of problems. Impressive performance gains were already shown in language translation tasks for the original transformer \cite{vasvani17}.
The recent public release of the transformer-based \ac*{LLM} ChatGPT by OpenAI has attracted major media attention \cite{radford_language_2019}. ChatGPT is a version of the \ac*{GPT} models developed by OpenAI \cite{radford_improving_2018,radford_language_2019}. Successively larger models (GPT: 110 million parameters, GPT-2: 1.5 billion parameters, GPT-3: 175 billion parameters \cite{floridi_gpt-3_2020}) have led to an impressive ability to generate human-like text. The new GPT-4 reportedly has 100 trillion parameters and works with both text and images \cite{openai23}. The \ac*{GPT} models use a decoder-only architecture with the decoder architecture relatively unchanged from the original transformer model\cite{lim_temporal_2021} \cite{vasvani17,radford_improving_2018,radford_language_2019}. The \ac*{GPT} models are trained on unlabeled data, allowing the models to be trained on data that require virtually no pre-processing, alleviating the need for manually labeled data.

The motivation for training a transformer neural network architecture stems from the successful applications of transformers in many fields. Examples include ``AlphaFold'', a neural network incorporating transformers for predicting a protein's three-dimensional structure based on the amino acid sequence with atomic accuracy \cite{eisenstein_artificial_2021,alquraishi_machine_2021,jumper_highly_2021}. Transformers have also been used for classification \cite{chen_generative_2020,dosovitskiy_image_2020}, object detection, image segmentation, and pose estimation \cite{han_survey_2023}. In \cite{chen_mesh_transformer_2022}, a transformer was adapted to have as input three-dimensional meshes and perform shape classification and segmentation tasks. Two recent publications use transformers to improve the estimation of counterfactual outcomes in, e.g., healthcare settings \cite{dedhia_scout_2022,melnychuk_causal_2022}. Furthermore, studies have shown that transformers or transformer-based models can be used for time series forecasting based on previously observed data, such as predicting the prevalence of influence \cite{wu_deep_2020,li_enhancing_2020,zhou_informer_2021}. In \cite{lim_temporal_2021}, a transformer-based time series forecasting model named \text{Temporal Fusion Transformer} was proposed for multi-horizon forecasting, i.e., prediction of multiple variables of interest. The model incorporates previous time series data, a mix of static and time-dependent inputs, and known future inputs.

Following the architecture of most sequence-to-sequence models, the transformer has an encoder-decoder structure \cite{vasvani17}. Given an input sequence $\bm{x}=\left( x_1, \ldots, x_n\right) $, the encoder generates an intermediate sequence $\bm{z}=\left( z_1, \ldots, z_n\right)$. The decoder uses $\bm{z}$ to element-wisely generate the output $\bm{y}=\left( y_1, \ldots, y_n\right)$. Previously generated output values are auto-regressively used as supplementary input \cite{vasvani17}.
Commonly, models for processing data with sequential dependencies use recurrent layers to process individual values sequentially \cite{aggarwal_neural_2018}. Transformers, in contrast, solely rely on so-called attention mechanisms in the encoder and decoder to represent dependencies between input and output sequences \cite{vasvani17,bahdanau_neural_2016}. Attention was initially introduced in \cite{bahdanau_neural_2016} to improve the performance of sequence models for long-range dependencies. The transformer neural network used a scaled dot-product attention mechanism given by \cite{vasvani17}:

	\begin{equation}
		\text{Attention}\left( \bm{Q},\bm{K},\bm{V}\right) = \text{softmax}\left( \frac{\bm{Q}\bm{K}^T}{\sqrt{d_k}}\right) \bm{V}
	\end{equation}

where $\bm{Q}$, $\bm{K}$, and $\bm{V}$ are vectors of dimension $d_k$, $d_k$, and $d_v$, named query, keys, and values, respectively, and softmax is the softmax activation function. In the transformer, a so-called multi-head attention was used instead of using the attention mechanism a single time. The attention function is performed in parallel $h$ times, where $h$ is a selected parameter (eight in \cite{vasvani17}) and represents the number of attention heads. Additionally, $\bm{Q}$, $\bm{K}$, and $\bm{V}$ are linearly projected to $d_k$, $d_k$, and $d_v$ dimensions, respectively, using projection matrices $\bm{W}$ that are learned during model training \cite{vasvani17}:

	\begin{equation}
		\text{head}_i = \text{Attention}\left( \bm{Q}\bm{W}_i^Q,\bm{K}\bm{W}_i^K,\bm{V}\bm{W}_i^V\right)\qquad\text{with}\quad i = 1, \ldots, h. 
	\end{equation}

This operation produces $d_v$-dimensional output for each attention head. The output of the $h$ attention heads is then concatenated and linearly projected again using the learned projection matrix $\bm{W}^O$ to produce the multi-head attention output \cite{vasvani17}:

	\begin{equation}
		\text{MultiHead}\left( \bm{Q},\bm{K},\bm{V}\right) =\text{Concat}\left( \text{head}_1, \ldots, \text{head}_h\right) \bm{W}^O.
	\end{equation}

From this, the dimensions of the projection matrices follow \cite{vasvani17}:

	\begin{equation}
		\begin{aligned}
			\bm{W}_i^Q & \in \mathbb{R}^{d_{model} \times d_k}\\
			\bm{W}_i^K & \in \mathbb{R}^{d_{model} \times d_k}\\
			\bm{W}_i^V & \in \mathbb{R}^{d_{model} \times d_v}\\
			\bm{W}^O & \in \mathbb{R}^{hd_v \times d_{model}},\\
		\end{aligned}
	\end{equation}

with the selected model size $d_{model}$ (e.g., $d_{model}=512$ in \cite{vasvani17}).

There are contradicting studies on whether attention can provide reasoning for a model's predictions by laying out the relative importance of inputs \cite{jain_attention_2019,wiegreffe_attention_2019}. In \cite{jain_attention_2019}, the authors compared attention weights with feature importance and found a low correlation, arguing that attention provides little transparency. In contrast, in \cite{wiegreffe_attention_2019} the authors argue that attention can provide a plausible explanation for a model's predictions, however, not ``\textit{one true, faithful interpretation of the link} $\ldots$ \textit{between inputs and outputs}''.

\begin{figure}[!htb]
	\centering
	\includegraphics[width=0.8\linewidth]{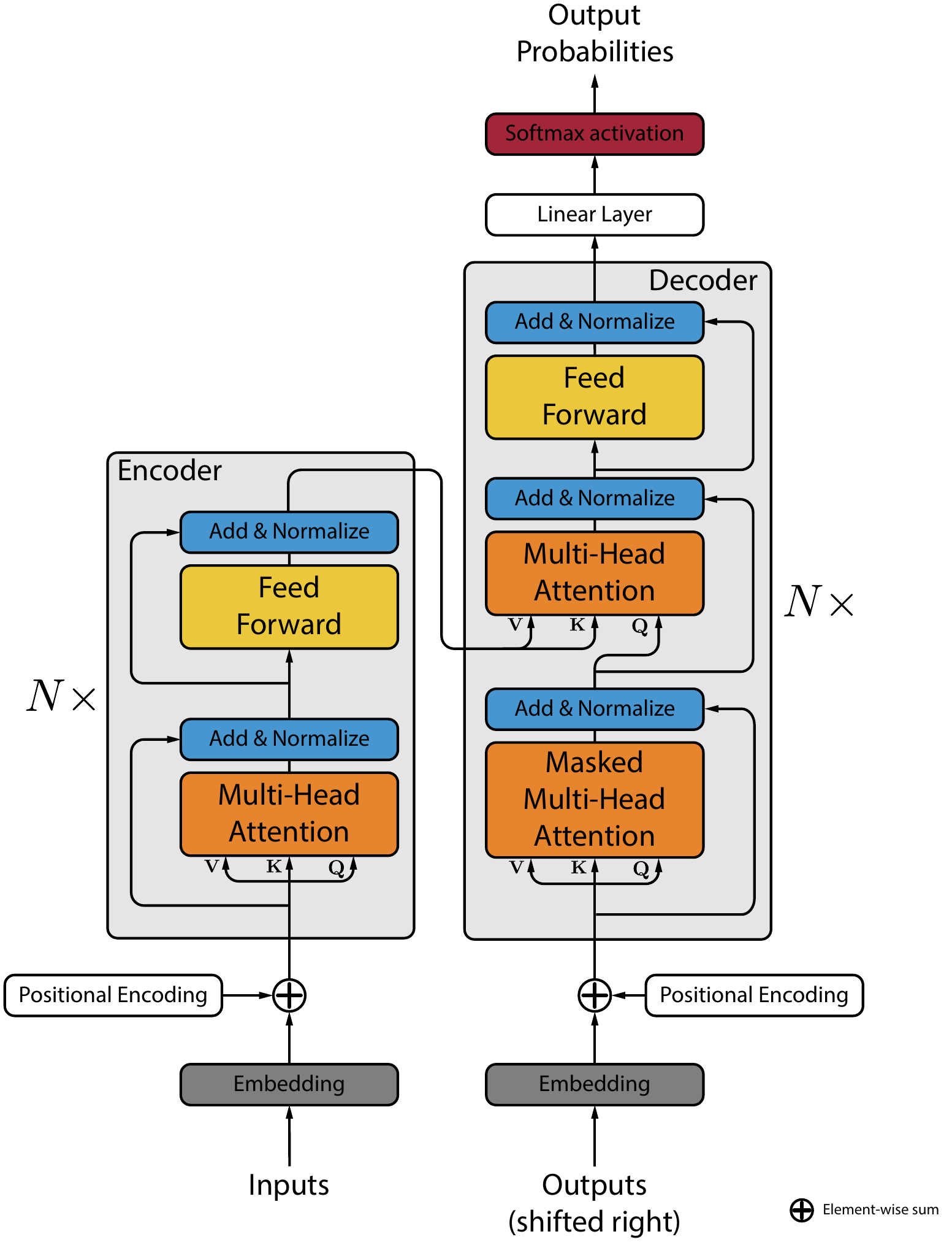}
	\caption{Architecture of the transformer model. (adapted from \cite{vasvani17})}
	\label{fig:transformer}
\end{figure}

Using the multi-head attention mechanism, the architecture of the transformer is depicted in Fig.~\ref{fig:transformer}.
The encoder layer consists of a multi-head attention mechanism and a simple fully connected feed-forward layer, both having residual connections \cite{vasvani17,he_deep_2015}. After both attention and feed-forward layers, normalization is performed. The encoder layer is repeated $N$ times ($N=6$ in \cite{vasvani17}). Keys, values, and queries of the multi-head attention mechanism are the embedded model input (first encoder layer) or the output of the previous encoder layers, making the attention mechanism a self-attention mechanism.

A similar architecture is used for the decoder layers, with an additional multi-head attention mechanism (``encoder-decoder attention'', middle sub-layer in Fig.~\ref{fig:transformer}) that uses as values and keys the output of the encoder and as queries, the output of the previous decoder layers \cite{vasvani17}. The encoder-decoder multi-head attention mechanism is similarly followed by layer normalization and includes residual connections. Furthermore, the self-attention mechanism attending to the embedded model output (first decoder sub-layer in Fig.~\ref{fig:transformer}) is masked and the output sequence is shifted to the right by one position. Masked self-attention effectively prevents the decoder from ``looking into the future'', i.e., the decoder predictions can only depend on known past output positions. Again, the decoder layer is repeated $N$ times. Output from the decoder is finally passed through a linear layer followed by softmax activation to obtain the output probabilities.

The inputs and outputs of the model have to be converted to vectors of dimension $d_{model}$ \cite{vasvani17}. In \ac*{NLP} applications, commonly learned embeddings are used for this purpose \cite{jurafsky_speech_2009}. 

Unlike \acp*{RNN}, the transformer does not use any recurrence and, therefore, does not directly have information about the location of a value in a sequence \cite{vasvani17}. Instead, positional information is injected by adding a positional encoding to the embedded input and output. The positional encoding consists of sine and cosine functions of different frequencies.

Additionally, transformers can be advantageous in multiple aspects compared to \acp*{RNN}: Long-range dependencies in time sequences are problematic due to the recurrent nature of \acp*{RNN} \cite{li_enhancing_2020,vasvani17}. Transformers seem to be better able to capture long-range dependencies because self-attention mechanisms connect all positions in a sequence with the same number of operations, compared to sequentially in \acp*{RNN} \cite{vasvani17}. Furthermore, \acp*{RNN} can suffer from exploding or vanishing gradients during training, making training the model more difficult \cite{aggarwal_neural_2018}.

\section{Generation of Training data}\label{sec:training_data}

The objective of this effort was to develop a transformer neural network surrogate to predict the non-linear and history-dependent response of composite microcells. Generation of the training dataset requires analysis of a microcell's response to deformation through scale transition from the micro-scale to the macro-scale. An introduction to \ac*{FEM}-based homogenization to solve the multi-scale \ac*{BVP} and generate the stress response of a microcell for finite-strain problems will be given in the following section (Sec.~\ref{sec:homogenization_intro}). 

\subsection{Solving the Multi-Scale BVP to Calculate the Microcell Stress Response}\label{sec:homogenization_intro}
\begin{figure}[!htb]
	\centering
	\includegraphics[width=0.6\linewidth]{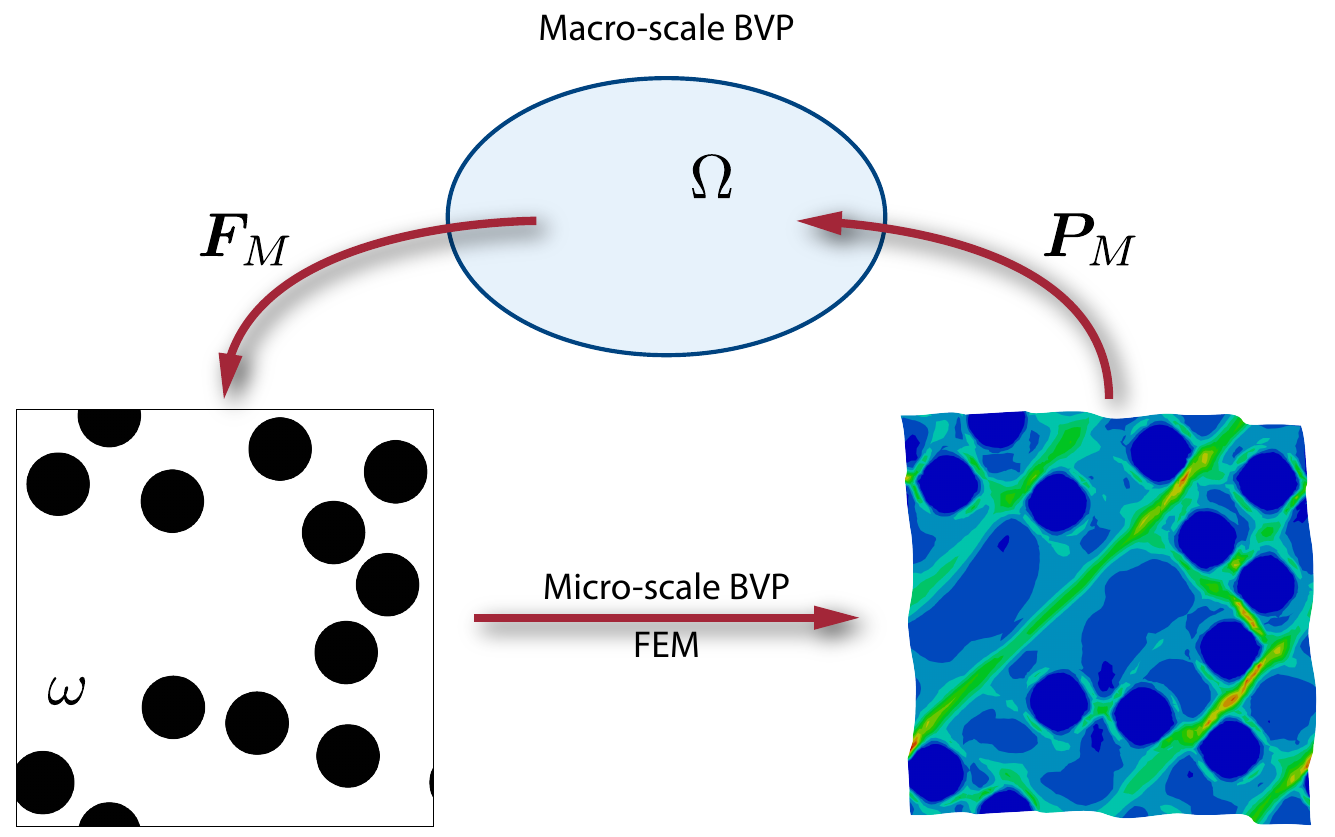}
	\caption{Scale-transition: Given a macro-scale deformation gradient $\bm{F}_M$ the corresponding homogenized Piola-Kirchhoff stress $\bm{P}_M$ is found by solving the micro-scale \ac*{BVP}. (adapted from \cite{wu20})}
	\label{fig:bvp}
\end{figure}

Given a (homogenized) macro-scale deformation gradient tensor $\bm{F}_M$, the homogenized Piola-Kirchhoff stress tensor of the underlying microcell $\bm{P}_M\left( \bm{X}, t\right)$ at (macro) material point $\bm{X}$ and time $t$ shall be found (see Fig.~\ref{fig:bvp}). The homogenized stress tensor $\bm{P}_M$ can be found by solving the micro-scale \ac*{BVP} using \ac*{FEM}. In the following equations, $M$ represents the macro-scale values and $m$ refers to values on the micro-scale. The given summary of the multi-scale homogenization problem is based on the derivation presented in \cite{wu20,wu_recurrent_2022}. 

The linear momentum equations at the macro-scale in the domain $\Omega$ and neglecting dynamic effects are given by \cite{wu20,wu_recurrent_2022}:

 \begin{equation}
	\bm{P}_M\left( \bm{X}\right) \cdot\nabla_0 + \bm{b}_M =\bm{0} \qquad \forall\bm{X}\in\Omega,
\end{equation}

with the gradient operator with regard to the reference configuration $\nabla_0$ and the body force $\bm{b}_M$. The linear momentum equations are subject to the Dirichlet boundary conditions on the boundary $\partial_D\Omega$ and Neumann boundary conditions on the boundary $\partial_N\Omega$:
\begin{align}
	\bm{u}_M\left( \bm{X}\right)&=\hat{\bm{u}}_M \qquad\forall\bm{X}\in\partial_D\Omega\\
	\bm{P}_M\left( \bm{X}\right) \cdot\bm{N}_M &=\hat{\bm{T}}_M\qquad\forall\bm{X}\in\partial_N\Omega,
\end{align}
with the displacements $\bm{u}_M$, the constrained displacements $\hat{\bm{u}}_M$ on $\partial_D\Omega$, the unit normal $\bm{N}_M$ on $\partial_N\Omega$, and the surface traction $\hat{\bm{T}}_M$.

On the micro-scale, the \ac*{BVP} is commonly defined on a parallelepipedic \ac*{RVE} with the boundaries of the \ac*{RVE} being planar faces $\partial\omega$. The linear momentum equations on the micro-scale are given by:

 \begin{equation}
	\bm{P}_m\left( \bm{x}\right) \cdot\nabla_0 =\bm{0} \qquad \forall\bm{x}\in\omega,
\end{equation}

subject to boundary conditions:

 \begin{equation}
	\bm{P}_m\left( \bm{x}\right) \cdot\bm{N}_m=\hat{\bm{T}}_m\qquad\forall\bm{x}\in\partial\omega,
\end{equation}

where the traction $\bm{T}_m$ acts on the boundary $\partial\omega$ with unit normal $\bm{N}_m$.

The macro scale deformation gradient tensor $\bm{F}_M$ and stress tensor $\bm{P}_M$ are given by volume average over $\omega$ of the micro scale deformation gradient $\bm{F}_m$ and micro scale stress tensor $\bm{P}_m$, respectively \cite{wu20}:
\begin{align}
	\bm{F}_M\left( \bm{X}, t\right) &=\frac{1}{V\left( \omega\right) }\int_\omega \bm{F}_m\left( \bm{x}, t\right) d\bm{x}\\
	\bm{P}_M\left( \bm{X}, t\right) &=\frac{1}{V\left( \omega\right) }\int_\omega \bm{P}_m\left( \bm{x}, t\right) d\bm{x}
\end{align}
Energy consistency requires the Hill-Mandel condition to be fulfilled \cite{wu20,wu_recurrent_2022,peric_micro-macro_2011}:

 \begin{equation}
	\bm{P}_M : \delta\bm{F}_M = \frac{1}{V\left( \omega\right) }\int_\omega \bm{P}_m : \delta\bm{F}_m d\bm{x}.
\end{equation}

On the micro-scale, the displacement field $\bm{u}_m$ with an introduced pertubation field $\bm{u}'\left( x\right)$, the macro-scale displacement gradient $\bm{F}_M$, and unit tensor $\bm{I}$ is defined by \cite{wu20}:

 \begin{equation}
	\bm{u}_m\left( \bm{x}\right) =\left( \bm{F}_M - \bm{I}\right) \cdot\left( \bm{x}-\bm{x}_0\right) + \bm{u}'\left( \bm{x}\right).
\end{equation}

The point $\bm{x}_0$ is a reference point in $\omega$.
Using trial and test functions, the weak form of the micro-scale \ac*{BVP} can be formulated:

 \begin{equation}
	\int_\omega\bm{P}_m\left( \bm{u}'\right) : \left( \partial\bm{u}'\otimes\nabla_0\right)d\bm{x} = 0, \qquad\forall \partial\bm{u}' \in \mathcal{U}\left( \omega\right) \subset\mathcal{U}^{\text{min}}\left( \omega\right),
\end{equation}

where $\partial\bm{u}'\in\mathcal{U}\left( \omega\right)$ is a test function in the admissible kinematic vector field $\mathcal{U}\left( \omega\right)$, $\bm{u}'\in\mathcal{U}\left( \omega\right) \subset\mathcal{U}^{\text{min}}\left( \omega\right) $ shall be found, and $\mathcal{U}^{\text{min}}\left( \omega\right)$ is the minimum kinematic vector field.

Given a prescribed macro-scale deformation gradient $\bm{F}_M$, the weak form can be solved using \ac*{FEM} to find the homogenized stress tensor $\bm{P}_M$. On opposite faces of the \ac*{RVE} $\bm{x}^+\in\omega^+$ and $\forall\bm{x}^-\in\omega^-$, \acp*{PBC} are applied:

 \begin{equation}
	\begin{aligned}
		\mathcal{U}^{\text{PBC}}\left( \omega\right) = &\left\lbrace \bm{u}'\in\mathcal{H}\left( \omega\right) | \bm{u}'\left( \bm{x^+}\right) = \bm{u}'\left( \bm{x^-}\right)\right. , \\
		&\left. \forall\bm{x}^+\in\omega^+ \quad\text{and corresponding}\quad	 \forall\bm{x}^-\in\omega^-\right\rbrace \subset \mathcal{U}^{\text{min}} \left( \omega\right) ,
	\end{aligned}
\end{equation}

with the Hilbert space $\mathcal{H}$. 
While various kinds of boundary conditions can be applied to fulfill the Hill-Mandel condition \cite{wu20}, \acp*{PBC} were chosen for this work for ease of implementation.

\subsection{Description of the Micro-Scale Composite Volume Elements}\label{sec:data_generation}
To generate the training dataset, a workflow was implemented to:
\begin{enumerate}
	\item Generate random composite \acp*{SVE}.
	\item Setup and execute \ac*{FEM} simulations with \acp*{PBC} and given homogenized load to generate the homogenized microcells' response by solving the micro-scale \acp*{BVP}.
	\item Save the microstructure data and homogenized loading and response (as sequences of strain increments and corresponding stress) in a database to use as training data.
\end{enumerate}
The term \ac*{SVE} refers to a volume element of the analyzed composite material that shows varying homogenized properties depending on the \ac*{SVE} realization \cite{wu19} (e.g., given multiple volume elements with similar \ac*{FVR}, each volume element will yield different results). This is in contrast to an \ac*{RVE}, which assumes statistical representativity.

\begin{figure}[!htb]
	\centering
	\begin{subfigure}{.25\textwidth}
		\centering
		\includegraphics[width=0.9\linewidth,frame]{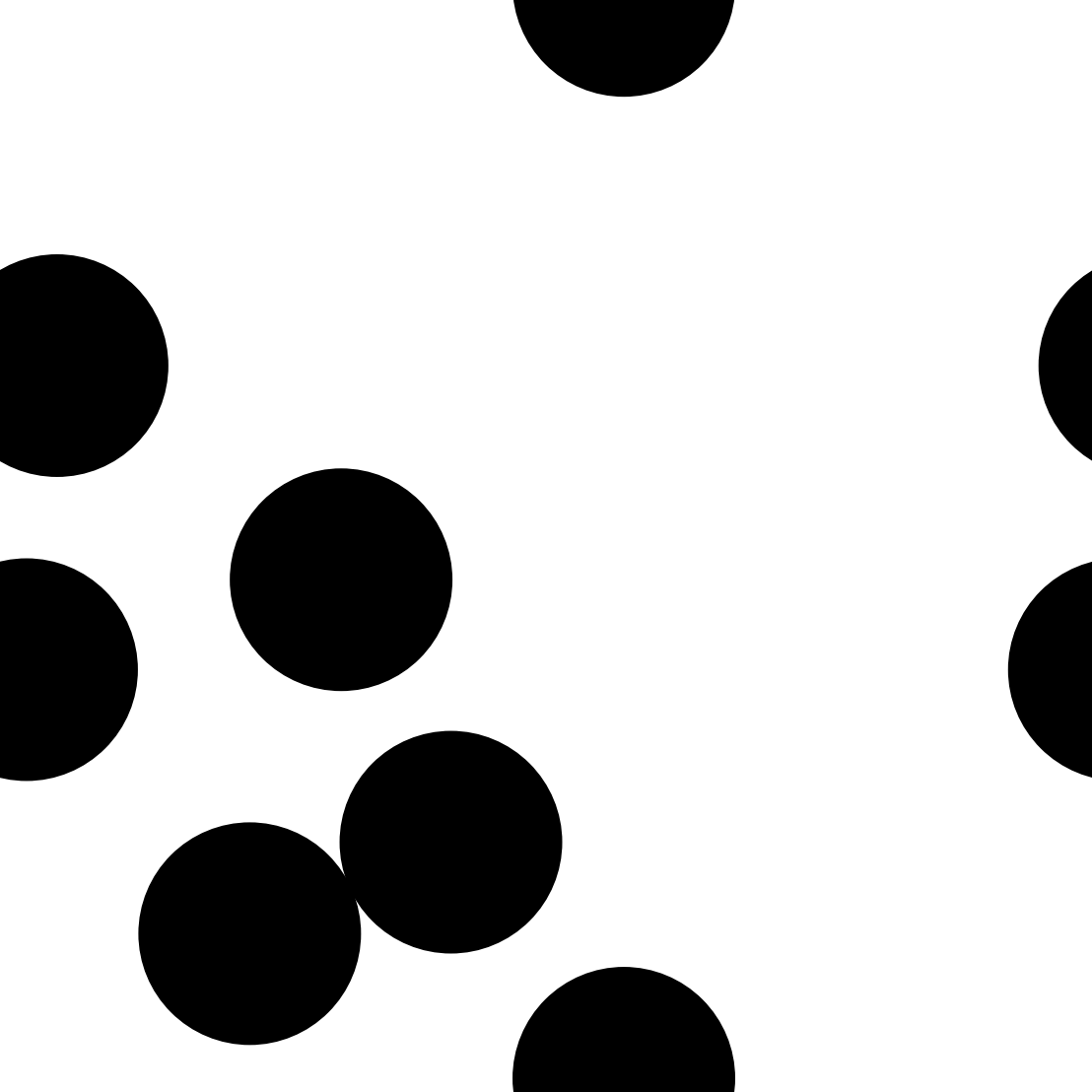}
		\caption{FVR=0.20.}
	\end{subfigure}%
	\begin{subfigure}{.25\textwidth}
		\centering
		\includegraphics[width=0.9\linewidth,frame]{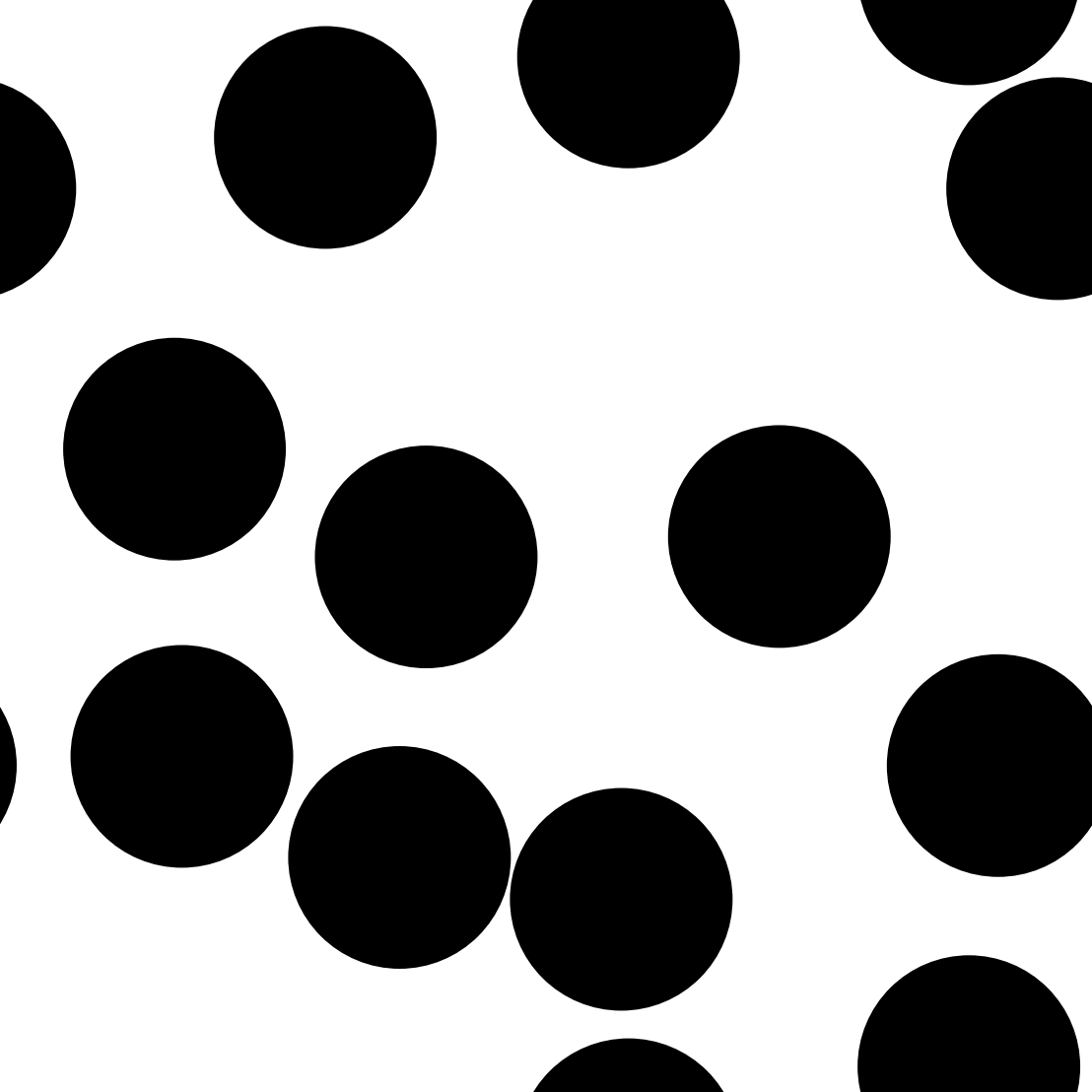}
		\caption{FVR=0.36.}
	\end{subfigure}
	\begin{subfigure}{.25\textwidth}
		\centering
		\includegraphics[width=0.9\linewidth,frame]{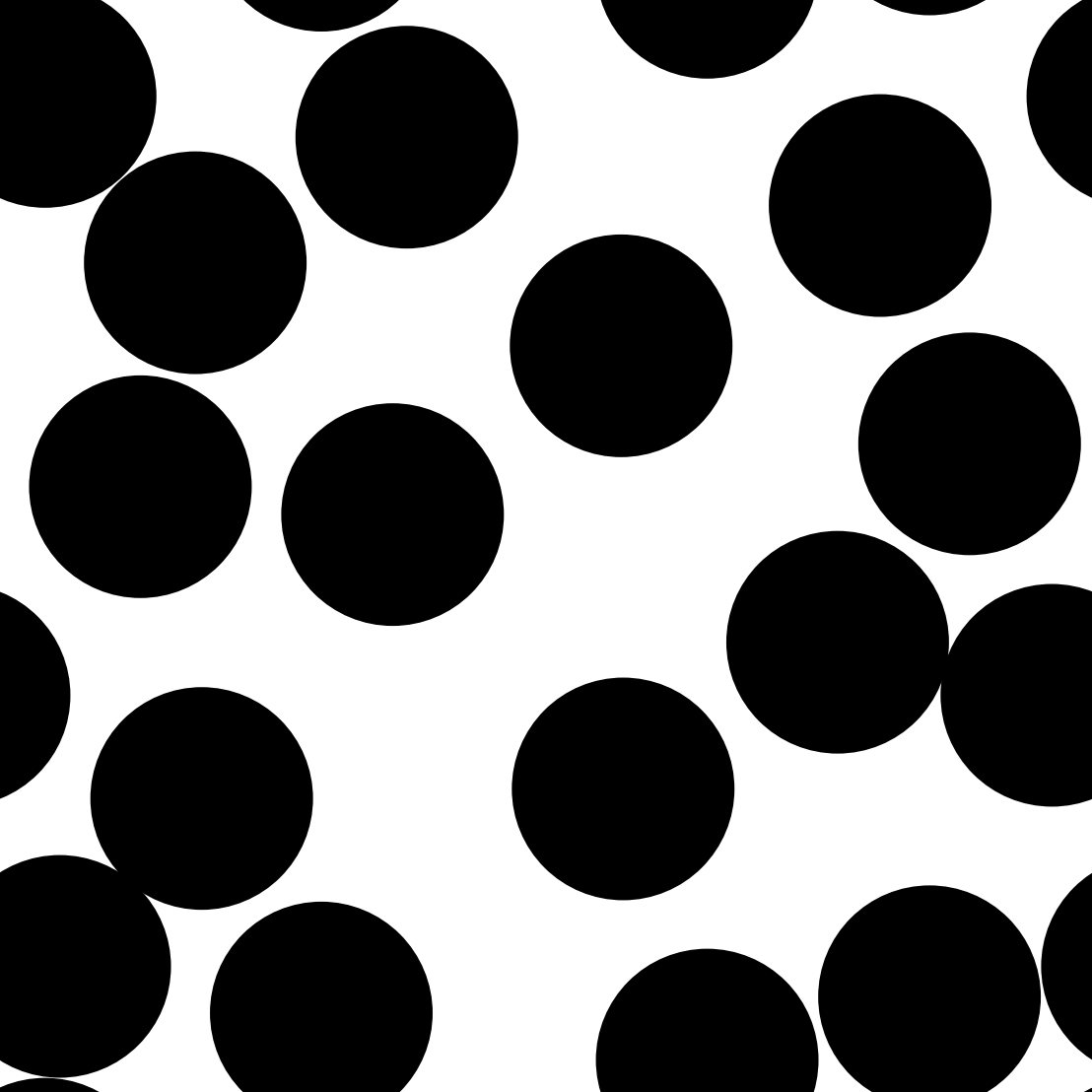}
		\caption{FVR=0.53.}
	\end{subfigure}
	\caption{Three random composite microcells with varying \acp*{FVR}.}
	\label{fig:training_figs}
\end{figure}

For simplicity, in this effort, the application was restricted to two-dimensional problems, more specifically to the transversal behavior of continuously reinforced composites assuming plane strain conditions. Therefore, the generated volume elements consisted of circular fibers inside a matrix (see Fig.~\ref{fig:training_figs} for three arbitrary \acp*{SVE} at different volume fractions). 
Given plane-strain conditions and with the $z$-axis being the out-of-plane direction (normal to an analyzed \ac*{SVE}), the three non-zero strain components $\bm{E}_{M,xx}$, $\bm{E}_{M,yy}$, and $\bm{E}_{M,xy}$ can be prescribed resulting in four stress components $\bm{S}_{M,xx}$, $\bm{S}_{M,yy}$, $\bm{S}_{M,zz}$, and $\bm{S}_{M,xy}$.

Periodic \acp*{SVE} were generated by randomly placing fibers inside the microcell until a desired \ac*{FVR} was reached. Uniform distributions were used to place the fibers. The uniformly distributed fiber locations inside the generated microstructures may not represent the fiber location distributions found in a composite material \cite{wu19}. Further analysis of the fiber location distributions could be an interesting objective for future model improvements. After generating the microcell geometry, the \ac*{SVE} was meshed with quadratic (to prevent locking effects) plain strain triangles using the Python API of the open-source software \textit{Gmsh} \cite{geuzaine_gmsh_2009}. Periodic nodes were generated on opposite sides of the \ac*{SVE} during meshing to allow \acp*{PBC} to be applied. The generated mesh was then used to generate an input file for the \ac*{FEM} solver \textit{CalculiX} \cite{dhont_calculix}. \acp*{PBC} were applied, mapping the macro-scale deformation gradient to the \ac*{SVE}. An incremental plasticity model with a multiplicative decomposition of the deformation gradient was used to model the matrix behavior. Linear elastic behavior was assumed until reaching the yield strength, followed by perfectly plastic behavior (accumulating plastic strain at constant stress with further loading) \cite{dhondt_finite_2004}. The matrix plasticity introduces a history-dependency of the microcell response. Fibers were modeled linear elastically. Using \textit{CalculiX}, the micro-scale \ac*{BVP} was solved and the strains and stresses at the integration points were recorded. Subsequently, the volume average strain and stress were calculated as the homogenized properties.

Following \cite{wu20, wu_recurrent_2022}, the generated training data consisted of both cyclic strain sequences and random strain walks. Strain sequences were generated with a maximum length of 100 strain increments. To apply boundary conditions to the micro-scale \ac*{BVP}, the deformation gradient tensor $\bm{F}_M$ was required. Therefore, loading paths, again following \cite{wu20}, were derived from a sequence of right stretch tensors $\bm{U}_{M_0}, \bm{U}_{M_1}, \ldots \bm{U}_{M_n}$. At each time step, a load increment $\Delta\bm{U}_{M_n}=\bm{U}_{M_n}-\bm{U}_{M_{n-1}}$ was applied. The increment of the right stretch tensor can be represented using orthogonal eigenvectors $\bm{n}_1$ and $\bm{n}_2$ (only the 2D-case is addressed here) that control the loading direction and eigenvalues $\Delta\lambda_1$ and $\Delta\lambda_2$, governing the load increment size \cite{wu20}:

 \begin{equation}
	\Delta\bm{U}_{M}=\Delta\lambda_1\bm{n}_1\otimes\bm{n}_1+\Delta\lambda_2\bm{n}_2\otimes\bm{n}_2.
\end{equation}

To generate the random loading paths, random orthogonal vectors were generated:

 \begin{equation}
	\begin{aligned}
		\bm{n}_1 &= \left[ \cos\alpha\quad\sin\alpha\right] ^T\\
		\bm{n}_2 &= \left[ -\sin\alpha\quad\cos\alpha\right] ^T,
	\end{aligned}
\end{equation}

with the angle $\alpha$ being randomly chosen from a uniform distribution with ${\alpha\in\left[ 0,\pi\right] }$. Corresponding eigenvalues were generated as:

 \begin{equation}
	\begin{aligned}
		\Delta\lambda_1&=\sqrt{\mathcal{R}}\cos\left( \theta\right) \\
		\Delta\lambda_2&=\sqrt{\mathcal{R}}\sin\left( \theta\right) ,
	\end{aligned}
\end{equation}

where $\mathcal{R}$ and $\theta$ are uniformly distributed variables in the intervals $\mathcal{R}\in\left[ \Delta R_{min}^2, \Delta R_{max}^2\right]$ and $\theta\in\left[ 0,2\pi\right] $. This guarantees the total step size to be bounded between $\Delta R_{min}$ and $\Delta R_{max}$:

 \begin{equation}
	\Delta R_{min} \le \sqrt{\Delta\lambda_1^2 + \Delta\lambda_2^2} \le \Delta R_{max}.
\end{equation}

A lower step size of $\Delta R_{min}=5\times10^{-5}$ and an upper limit of $\Delta R_{max}=2.5\times 10^{-3}$ were used for random strain sequences. Furthermore, the maximum total stretch change was limited to 0.05 for each component of the stretch tensor.

Additional to random loading paths, cyclic loading paths were generated, randomly choosing between uniaxial loading in $x$ or $y$ direction, biaxial loading ($x$ and $y$ loading), or pure shear loading. To generate cyclic strain sequences, the lower step size was increased to $\Delta R_{min}=5\times10^{-4}$.

To define the boundary conditions using \acp*{PBC}, the deformation gradient was calculated from the right stretch tensor:

 \begin{equation}
	\bm{F}_M=\bm{R}_M\cdot\bm{U}_M,
\end{equation}

where $\bm{R}_M$ is a rotation tensor set to $\bm{R}_M=\bm{I}$ in this effort due to the frame indifference of the micro-scale \ac*{BVP} \cite{wu20}. The Green-Lagrange strain tensor $\bm{E}_M$ as input to the \ac*{NN} surrogate was then calculated:

 \begin{equation}
	\bm{E}_M=\frac{1}{2}\left( \bm{U}_M^2-\bm{I}\right). 
\end{equation}

Corresponding to the homogenized Green-Lagrange strain tensor $\bm{E}_M$, the transformer \ac{NN} surrogate was trained and tested using the homogenized second Piola-Kirchhoff stress $\bm{S}_{M}$, obtained from the solution of the micro-scale \ac*{BVP} using \cite{wu20}:
\begin{equation}
	\bm{S}_{M} = \bm{F}_M^{-1}\cdot\bm{P}_M.
\end{equation}

Following the outlined data generation procedure, the training data for the \ac*{NN} consisted of sequences of strain tensors $\bm{E}_{M,1}, \bm{E}_{M,2}, \ldots, \bm{E}_{M,t}$ and corresponding stress tensors $\bm{S}_{M,1}, \bm{S}_{M,2}, \ldots, \bm{S}_{M,t}$. While a stress update algorithm was required when generating the training data using \ac*{FEM}, the surrogate model was trained to directly predict the next total stress based on the strain history and the previous stress state, which was stored as an internal state.
The transformer learned the nonlinear strain-stress behavior over several strain increments from the ground truth data.

\begin{figure}[!htb]
	\centering
	\begin{subfigure}{.6\linewidth}
		\centering
		\includegraphics{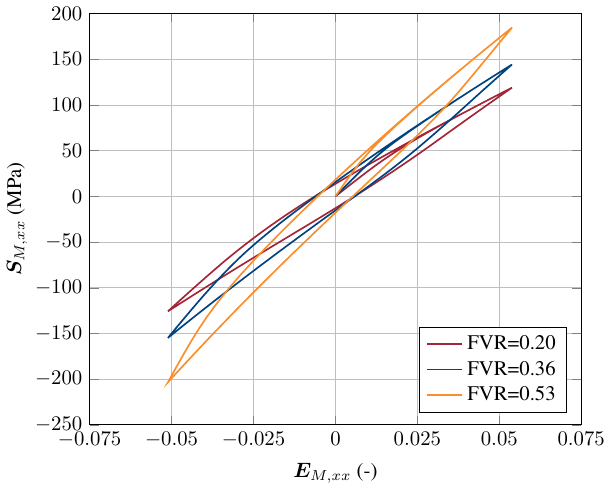}
		\caption{Cyclic strain sequence.}
		\label{fig:cyc_train}
	\end{subfigure}%
	\vspace{0.1mm}
	\begin{subfigure}{.6\linewidth}
		\centering
		\includegraphics{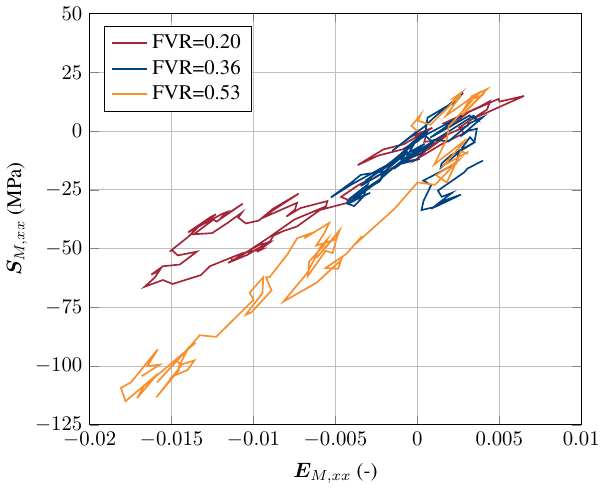}
		\caption{Random strain sequence.}
		\label{fig:rand_train}
	\end{subfigure}%
	\caption{Homogenized stress-strain response of three random \ac*{SVE} realizations with different \acp*{FVR} for cyclic (\ref{fig:cyc_train}) and random (\ref{fig:rand_train}) loading.}
	\label{fig:training_curves}
\end{figure}

The homogenized response of three random microcells with different \acp*{FVR} to cyclic loading (Fig.\ref{fig:cyc_train}) and random loading (Fig.\ref{fig:rand_train}) can be observed in Fig.\ref{fig:training_curves}. In Fig.\ref{fig:cyc_train}, a clear influence of the \ac*{FVR} on the homogenized stiffness and strength is identified when the microcells are subjected to cyclic loading, and the hysteresis loop due to plastic deformation in the matrix phase is evident. Furthermore, applying random strain walks to the microcells (Fig.~\ref{fig:rand_train}) results in many different loading and unloading paths. This variety enables the transformer surrogate model to learn the homogenized composite response across a wide range of unloading scenarios.

\section{Dimensionality Reduction of Microstructure Representation}\label{sec:microstructure_representation}
Homogenization of heterogeneous materials requires knowledge about the composition and internal structure of the material. To represent the microstructure a set of descriptors is required that can capture the features governing material behavior while being sufficiently low-dimensional to be tractable \cite{xu_machine_2015,cecen_versatile_2016}. Common examples of microstructure descriptors include volume fraction, inclusion size, or inclusion spacing. In \cite{mozaffar19}, \ac*{FVR}, fiber radius, and mean fiber distance were used to train \acp*{RNN} to predict the homogenized response of continuously reinforced composites. However, selecting microstructure descriptors intuitively might not sufficiently capture information and, when examining different types of microstructures, might not be generally applicable. Instead, an adaptive microstructure encoding that applies to a wide range of microstructures without manual feature selection is preferable. 

It has been shown that the relative spatial distribution of local heterogeneity based on two-point statistics can be used to effectively capture microstructure features \cite{brough_materials_2017,cecen_versatile_2016,ford_machine_2021}. Two approaches were be examined in this effort: a microstructure representation based on dimensionality-reduced two-point statistics and a learned microstructure auto-encoding where a \ac*{CNN} reduces the dimensionality and extracts information from an image of the microstructure. An introduction to two-point statistics will be given first.

\begin{figure}[!htb]
	\centering
	\includegraphics{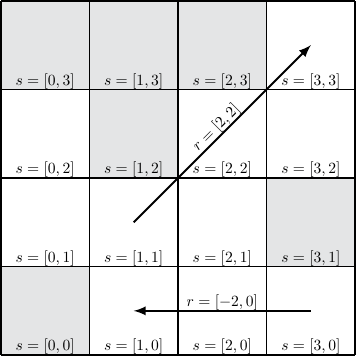}
	\caption{Schematic depiction of two-point correlations for a simple discretized microstructure. (adapted from \cite{brough_materials_2017})}
	\label{fig:two_point_stat}
\end{figure}

\begin{figure}[!htb]
	\centering
	\begin{subfigure}{.3\textwidth}
		\centering
		\includegraphics[width=1\linewidth]{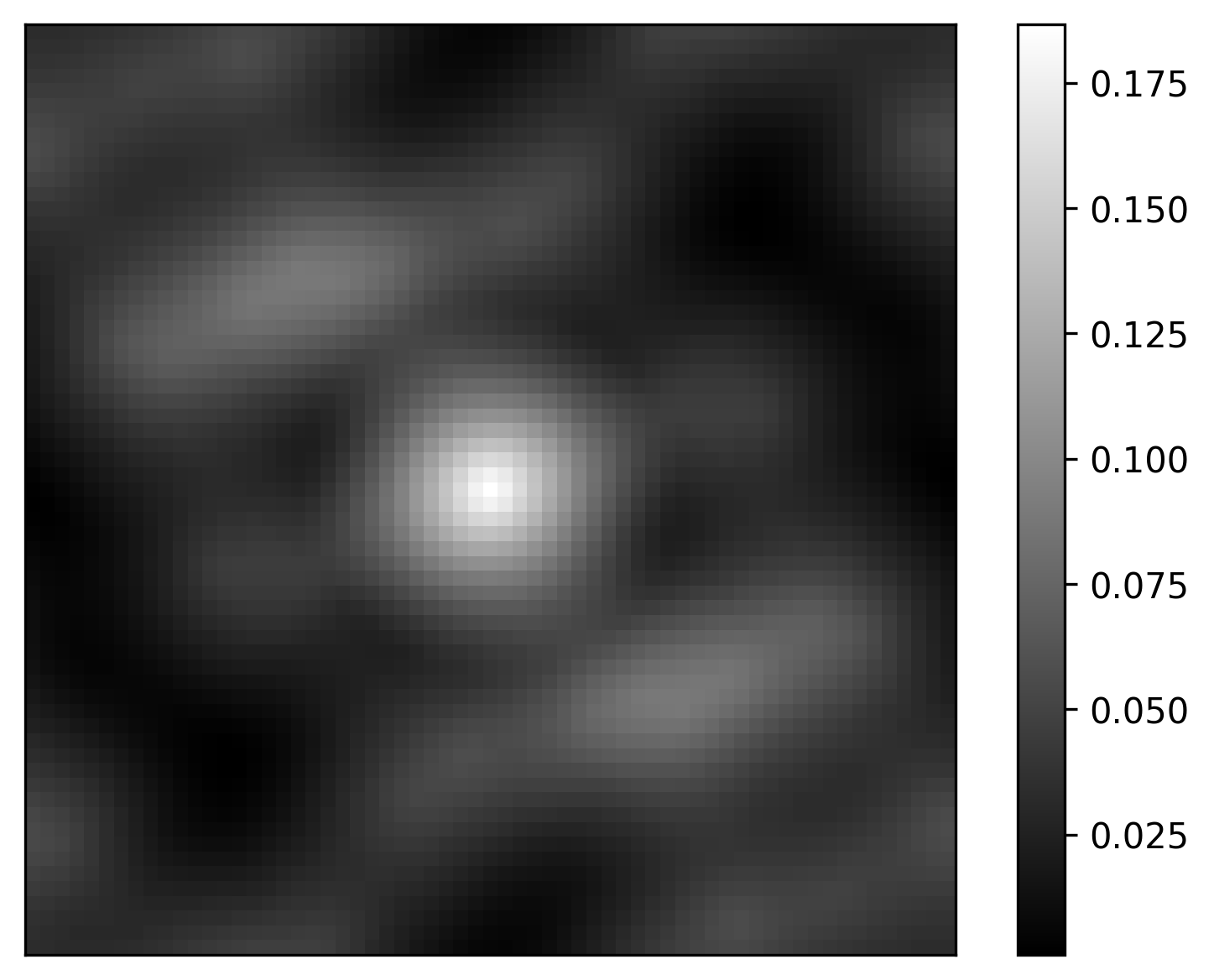}
		\caption{FVR=0.20.}
		\label{auto_corr_micro-fvr0d2}
	\end{subfigure}%
	\begin{subfigure}{.3\textwidth}
		\centering
		\includegraphics[width=1\linewidth]{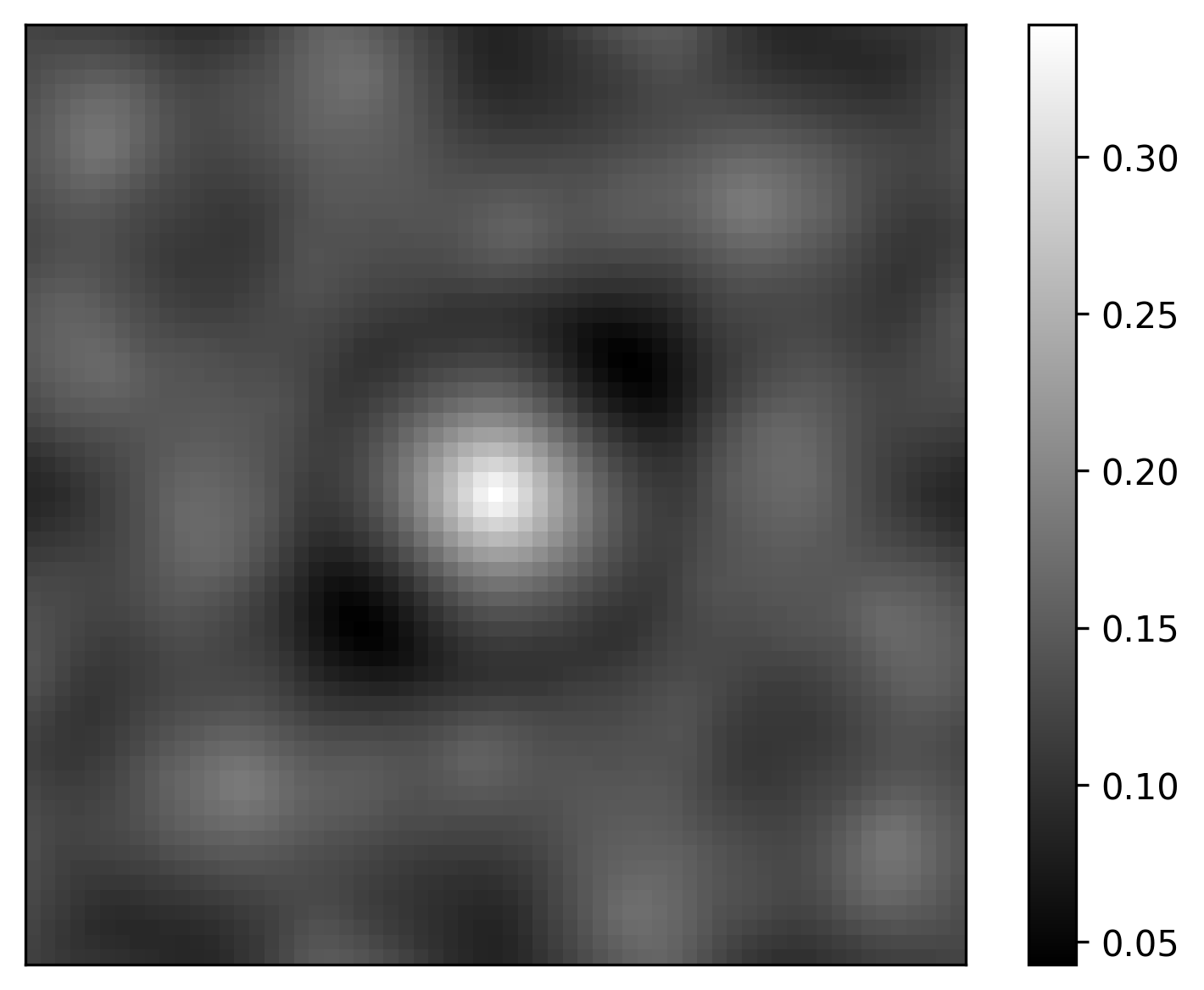}
		\caption{FVR=0.36.}
	\end{subfigure}
	\begin{subfigure}{.3\textwidth}
		\centering
		\includegraphics[width=1\linewidth]{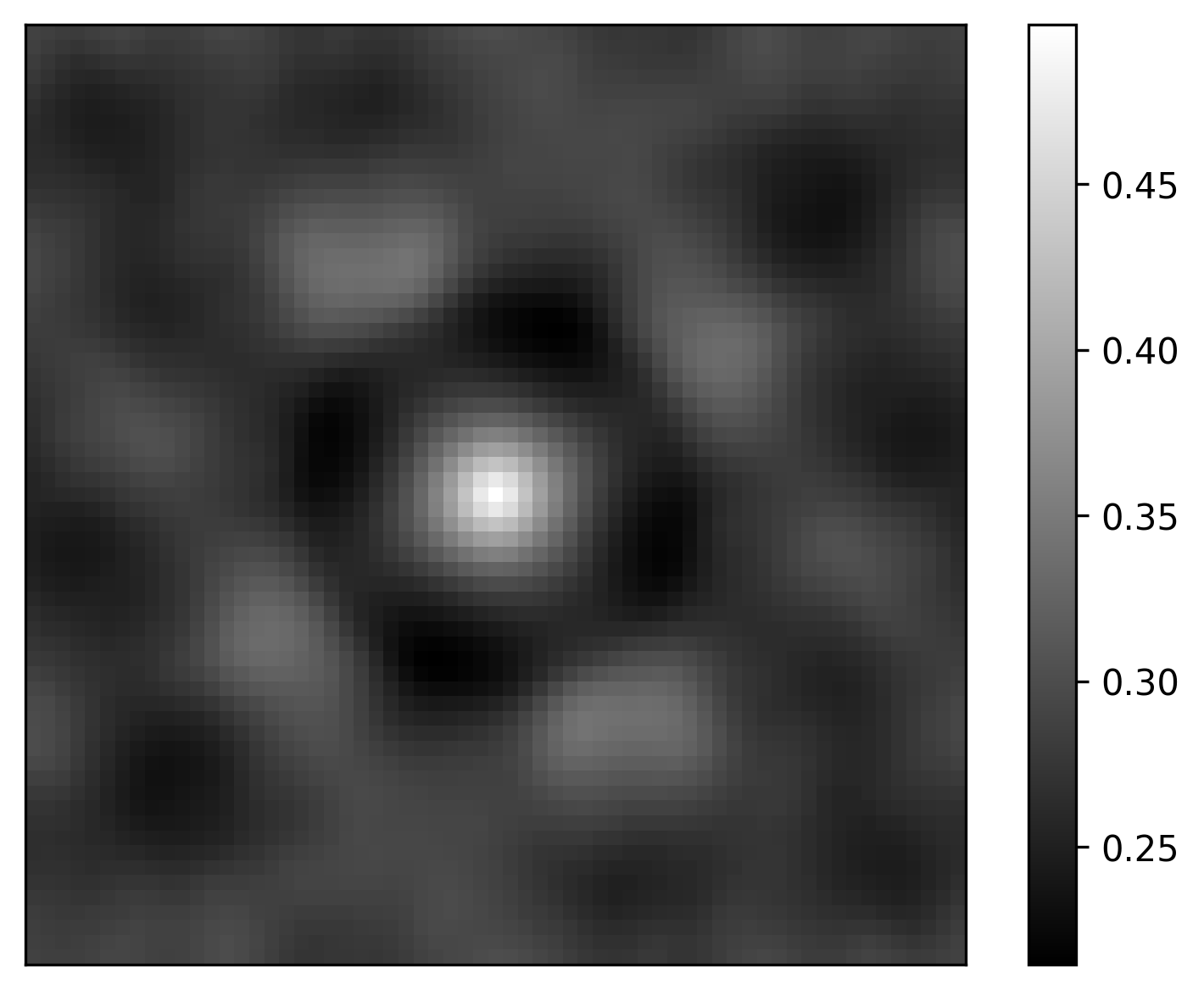}
		\caption{FVR=0.53.}
	\end{subfigure}
	\caption{Matrix phase auto-correlation corresponding to the microstructures shown in Fig.~\ref{fig:training_figs}.}
	\label{fig:auto_corr_micro}
\end{figure}

Two-point statistics or two-point spatial correlations can give information about the distribution of local states and the fractions of local states in a microstructure \cite{brough_materials_2017,brough_pymks_online,ford_machine_2021}. Effectively, two-point spatial correlations provide the probability that the start and end points of a vector are on a certain specified local state, respectively. A discretized (e.g., voxelized) microstructure $j$ is described by a microstructure function $m_j\left[ h;s\right] $ that gives the probability distribution for a local state $h\in H$, where $H$ are all possible local states, at each position $s\in S$ with the complete set of all positions $S$ \cite{brough_materials_2017}. In a voxelized two-dimensional microstructure with uniform cell size the position $s$ can be simply given by indices $i$ and $j$ of the voxels. A set of two-point correlations $f_j\left[ h, h'; r\right] $, providing relative spatial information of the local states $h$ and $h'$, can be calculated from the correlation of the microstructure function:

 \begin{equation}
	f_j\left[ h, h'; r\right] =\frac{1}{\Omega_j\left[ r\right] }\sum_{s} m_j\left[ h;s\right] m_j\left[ h';s+r\right],
\end{equation}

with a discrete vector $r$ inside of the microstructure domain and a normalization factor $\Omega_j\left[ r\right]$ dependent on $r$.
A simple two-phase microstructure (gray and white cells representing the different phases) with the positions $s$ is shown in Fig.~\ref{fig:two_point_stat}. The two-point statistics can be understood as the likelihood of encountering the first and second phase ($h$ and $h'$, respectively) of the material at the end (tail) and beginning (head) of a randomly placed vector $r$, respectively \cite{brough_materials_2017}. The matrix-phase auto-correlations displayed in Fig.~\ref{fig:auto_corr_micro} were generated using microstructures selected from the training set, which are depicted in Fig.~\ref{fig:training_figs}. The grayscale intensity values shown in Fig.~\ref{fig:auto_corr_micro} represent the probability of finding two points within the material that are both within the matrix phase. The two-point correlations, particularly for \ac*{FVR}=0.2 (Fig.~\ref{auto_corr_micro-fvr0d2}), display anisotropy, suggesting directional dependence of the phases within the material.


In this effort, a computational implementation of two-point statistics in the Python package \textit{\ac*{PyMKS}} \cite{brough_materials_2017} was used.

To reduce the dimensionality of the microstructure representation provided by two-point statistics, \ac*{PCA} was used \cite{hotelling_analysis_1933}. Two-point statistics generate a high-dimensional feature space that can contain redundant information, therefore \ac*{PCA} can be applied to efficiently remove redundant information and reduce the dimensionality \cite{brough_materials_2017}. Using \ac*{PCA}, the vector $\bm{f}_j\left[ l\right]$ containing all possible combinations of $h$, $h'$, and $r$ in $f_j\left[ h, h'; r\right]$, can be approximated as:

 \begin{equation}\label{eq:pca}
	\bm{f}_j\left[ l\right] \approx \overline{\bm{f}\left[ l\right] } + \sum_{n=1}^k \mu_{j,n} \bm{\phi}_n\left[ l\right].
\end{equation}

In Eq.~\ref{eq:pca}, $\overline{\bm{f}\left[ l\right] } $ represent the average values of all microstructures $j$ from the calibration data for all $l$, $ \mu_{j,n} $ are the principal component scores (effectively reduced-dimensionality microstructure descriptors), $\bm{\phi}_n\left[ l\right]$ are orthonormal vectors (principal components) that retain the maximum variance under projection, and $k$ is the number of selected components \cite{hotelling_analysis_1933,brough_materials_2017,tipping_mixtures_1999}. The principal component scores $\mu_{j,n}$ are ordered by their contributions to the variances, with $n=1$ having the highest contribution.

\begin{figure}[!htb]
	\centering
	\includegraphics{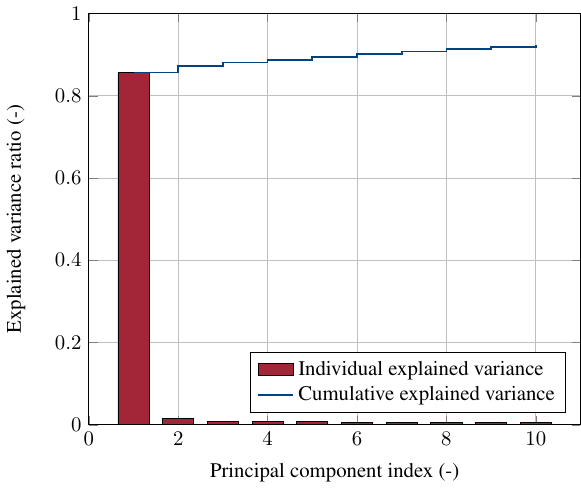}
	\caption{Contribution of the principal components to the variance for a random set of 1000 microstructures.}
	\label{fig:explained_variance}
\end{figure}

Randomly choosing 1000 microstructures from the training dataset, calculating two-point statistics and performing \ac*{PCA}, the individual and cumulative contribution to the variance of the first ten principal components are plotted in Fig.~\ref{fig:explained_variance}. The first principal component already explains more than 85\,\% of the dataset variance, while contributions of further components are quickly decreasing. Therefore, with six components explaining more than 90\,\% of the variance, \ac*{PCA} combined with two-point statistics is a highly efficient method to reduce the dimensionality of the microstructure description (compared to e.g., a geometric representation of the microstructure).

\begin{figure}[!htb]
	\centering
	\includegraphics{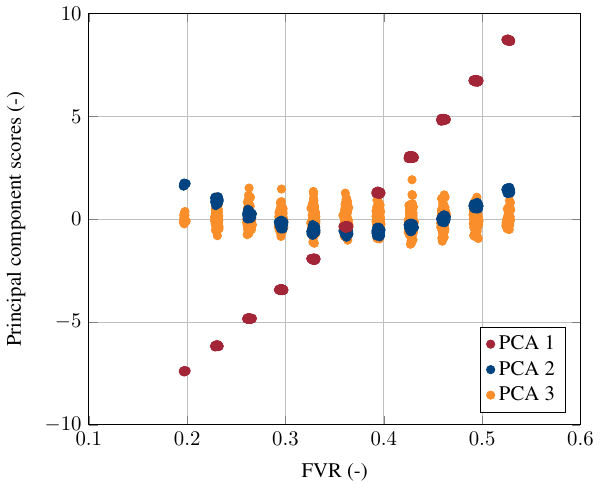}
	\caption{Analysis of principal component scores against the \ac*{FVR} for 1000 random microstructures.}
	\label{fig:pca_fvr}
\end{figure}

In continuously reinforced composites the predominant microstructure property is the ratio between fiber and matrix content \cite{ford_machine_2021}. To gain insight into the meaning of the principal component scores in microstructure representation, the scores of the first three principal components for 1000 random microstructures are plotted against the \ac*{FVR} in Fig.~\ref{fig:pca_fvr}.
A clear correlation between the first principal component score and the \ac*{FVR} is evident, with all microstructures with the same \ac*{FVR} having similar scores. This suggests that the first \ac*{PCA} score is a direct reflection of the \ac*{FVR}, unrelated to other microstructural characteristics. Again, similar values for the second \ac*{PCA} score are observed in microstructures with similar \acp*{FVR}. A low score is present for intermediate \ac*{FVR} and both high and low \ac*{FVR} exhibit high scores. However, the third \ac*{PCA} score exhibits variability in microstructures with similar \ac*{FVR}, implying that the \ac*{FVR} is not the dominant factor. Further examination is needed to identify the microstructural parameters represented by the principal component scores. Based on diminishing returns of later principal components, the surrogate model was trained using the first 3 principal components to describe the microstructure. These 3 components accounted for 88\,\% of the dataset variance.

It should be noted that a set of parameters describing the microstructure (such as \ac*{FVR}) might not necessarily capture features that govern strength behavior. As an example, composite stiffness can generally be well predicted purely based on \ac*{FVR}, while strength is generally highly dependent on local material inhomogeneity \cite{barnett_prediction_2021}. The validity of the chosen microstructure representation parameters has, therefore, to be evaluated with respect to their expressiveness for both stiffness and strength prediction. Since studies have indicated that the fiber spacing has a major impact on the transverse composite strength \cite{yang13}, future work will entail generating microstructure with controlled fiber spacing to analyze the impact of these microstructural features on the homogenized yield strength and the transformer's ability to capture this dependency.

As an alternative to two-point statistics with \ac*{PCA} for dimensionality reduction, a learned microstructure encoding was implemented. An image of the microstructure (as seen in Fig.~\ref{fig:training_figs}) was used as input to a \ac*{CNN}. The \ac*{CNN} successfully reduced the dimensionality of the microstructure image to a pre-defined number of parameters (e.g., 3 parameters) and created a hidden state that was used as input to the transformer model. Weights of the \ac*{CNN} were trained simultaneously with the transformer network to minimize the model prediction error. 

The intended purpose of the microstructure encoding learned by the \ac*{CNN} was to enable the automatic selection of microstructural features that control the homogeneous stiffness and strength response. An advantage of a learned microstructure encoding over two-point correlation could be that the \ac*{CNN} better adapts to features governing composite strength. In contrast, dimensionality-reduced two-point statistics provide pre-defined parameters that are not necessarily optimal to capture property governing features.

The used \ac*{CNN} architecture was roughly based on the \text{AlexNet} architecture \cite{krizhevsky17}. Five convolutional blocks were used with a progressively increasing number of output channels produced by the convolution. Maximum pooling layers were added in the second and fifth convolutional blocks for dimensionality reduction and batch normalization was used in the first, second, and fifth blocks to make training more stable. The convolutional blocks were followed by a three-layer feed-forward neural network producing the desired number of outputs. Dropout layers were used after the feed-forward layers to prevent overfitting. Throughout the \ac*{CNN}, \ac*{ReLu} activation functions were used. Black and white microstructure images that were input to the \ac*{CNN} were scaled to 32$\times$32 pixels. An overview of the \ac*{CNN} network architecture can be seen in Tab.~\ref{tab:CNN_architecture}. To assess the efficacy of the \ac*{CNN} in encoding microstructure information using dimensionality-reduced two-point statistics, an equal number of parameters were employed to characterize the microstructure, resulting in a \ac*{CNN} with 3 output parameters.
\begin{table}[!htb]
	\centering
	\caption{\ac*{CNN} architecture used to extract microstructure information from images.}
	\label{tab:CNN_architecture}
	\begin{tabular}{cm{7cm}cc}
		\toprule
		Layer nr & \multicolumn{1}{c}{Type and nonlinearity} & Input size & Output size \\\midrule
		& Input & & 32 $\times$ 32 $\times$ 1\\
		1 & Convolution (5$\times$5, 16 filters), Batch normalization, ReLu & 32 $\times$ 32 $\times$ 1 & 32 $\times$ 32 $\times$ 16\\
		2 & Convolution (3$\times$3, 32 filters), Batch normalization, ReLu, Max Pooling (2$\times$2) & 32 $\times$ 32 $\times$ 16 & 16 $\times$ 16 $\times$ 32\\
		3 & Convolution (3$\times$3, 64 filters), Batch normalization, ReLu & 16 $\times$ 16 $\times$ 32 & 16 $\times$ 16 $\times$ 64\\
		4 & Convolution (3$\times$3, 64 filters), Batch normalization, ReLu & 16 $\times$ 16 $\times$ 64 & 16 $\times$ 16 $\times$ 64\\
		5 & Convolution (3$\times$3, 128 filters), Batch normalization, ReLu, Max Pooling (2$\times$2) & 16 $\times$ 16 $\times$ 64 & 8 $\times$ 8 $\times$ 128\\
		6 & Flatten & 8 $\times$ 8 $\times$ 128 & 8192 \\
		7 & Fully connected, ReLu, Dropout (0.3) & 8192 & 2048 \\
		8 & Fully connected, ReLu, Dropout (0.3) & 2048 & 1024 \\
		9 & Fully connected, Batch normalization & 1024 & 3 \\
		\bottomrule
	\end{tabular}
\end{table}

\begin{figure}[!htb]
	\centering
	\includegraphics{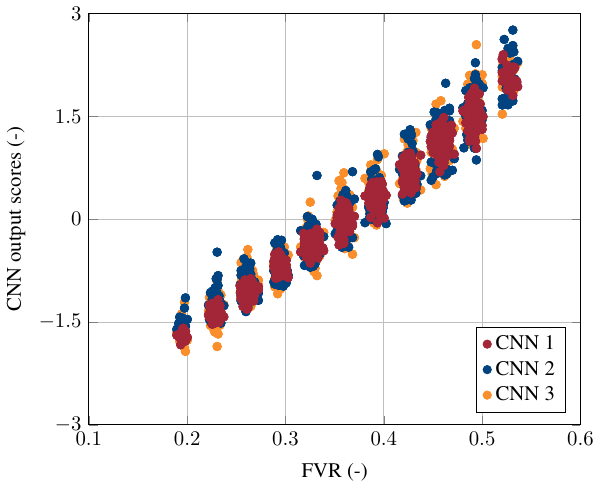}
	\caption{Output values of the trained \ac*{CNN} for 1000 random microstructures with varying \ac*{FVR}.}
	\label{fig:cnn_fvr}
\end{figure}

To analyze microstructure features extracted by the \ac*{CNN}, the output values of the trained network are plotted against the \ac*{FVR} in Fig.~\ref{fig:cnn_fvr} for 1000 random microstructures. All three output values show a clear correlation to the \ac*{FVR} with larger output values for larger \acp*{FVR}. The observed clear correlation shows that the \ac*{CNN} was able to extract governing microstructure features. However, because all output values were related to the \ac*{FVR}, likely a single output value would have been sufficient for the used training data. Introducing more variability into the microstructure, such as varying microcell size and fiber diameter, non-constant fiber diameter, or differently shaped inclusions, will require a larger number of parameters to describe the microstructure features.

\section{Transformer Surrogate Model and Training Procedure}\label{sec:surrogate}

In this effort, a transformer network was chosen as a surrogate to predict the history-dependent response of composite materials. While the original transformer \cite{vasvani17} consisted of an encoder-decoder architecture, only the decoder was used to build the surrogate model, similar to the \ac*{GPT} models \cite{radford_improving_2018}. The decoder's masked self-attention (see Sec.~\ref{sec:transformer_intro}) prevents the network from seeing ``future'', i.e., yet unknown, strain values in the input sequence. This guarantees that the predicted stress is only a function of the deformation history up to the current strain increment.

\begin{figure}[!htb]
	\centering
	\includegraphics[width=0.6\linewidth]{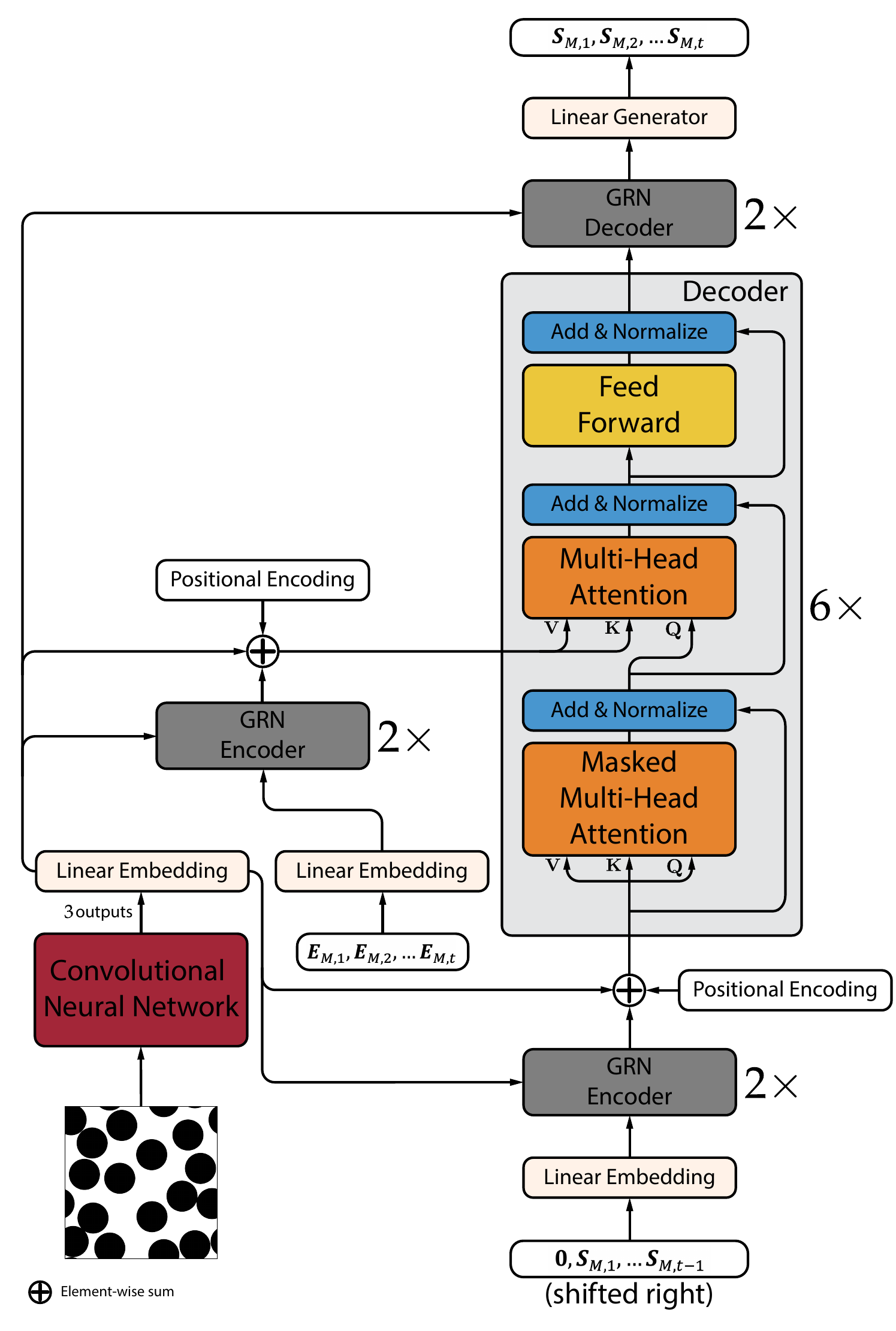}
	\caption{Architecture of the transformer network to predict the history-dependent response of heterogeneous composites using a \ac*{CNN} to extract features from microstructure images.}
	\label{fig:stress_transformer}
\end{figure}

\begin{figure}[!htb]
	\centering
	\includegraphics[width=0.3\linewidth]{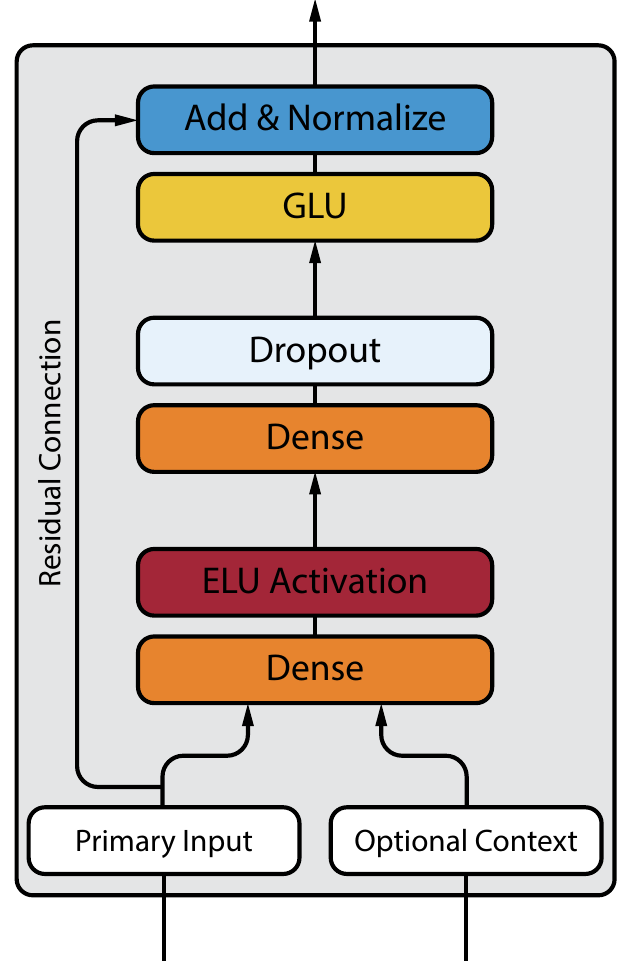}
	\caption{Architecture of the \ac*{GRN}. (adapted from \cite{lim_temporal_2021})}
	\label{fig:grn}
\end{figure}

An overview of the neural network architecture used in this effort is shown in Fig.~\ref{fig:stress_transformer}.
Transformers require an embedding to convert the input and the target sequence to vectors matching the transformer dimension. In the input path, the strain sequence, stress sequence, and the vector containing the microstructure information (either \ac*{PCA} scores or outputs from \ac*{CNN}) were linearly transformed to match the transformer dimension $d_{model}$. Two stacked \acp*{GRN} were used as encoders and decoder to embed static (time-independent) microstructure parameters. The \ac*{GRN} was adapted from the \textit{Temporal Fusion Transformer} \cite{lim_temporal_2021}. In addition to the primary input (strain or stress sequence), the \ac*{GRN} has an optional context vector (in our case the microstructure information) as input. By using gating mechanisms, the \ac*{GRN} can filter out insignificant inputs and can skip the \ac*{GRN} entirely if not relevant to the output \cite{lim_temporal_2021}. 
A schematic representation of the \ac*{GRN} architecture is shown in Fig.~\ref{fig:grn}. After passing through a linear layer, input and context are added and \ac*{ELU} activation \cite{clevert_fast_2016} function is applied. This is followed by a second linear layer, dropout, and a \ac*{GLU} \cite{dauphin_language_17} gating mechanism:

 \begin{equation}
	\text{GLU}\left( \bm{\gamma}\right) = \sigma\left( \bm{W}_{1} \bm{\gamma} +\bm{b}_{1} \right) \odot \left( \bm{W}_{2} \bm{\gamma} +\bm{b}_{2} \right) ,
\end{equation}

where $\bm{\gamma}$ is the input to the \ac*{GLU}, $\bm{W}_{1}$, $ \bm{W}_{2}$, $\bm{b}_{1}$, and $\bm{b}_{2}$ are the weights and biases, respectively, $\sigma\left( \cdot\right) $ is the sigmoid activation function, and $\odot$ the element-wise Hadamard product \cite{lim_temporal_2021}. A residual connection adds the \ac*{GRN} input and output, allowing the \ac*{GRN} to be skipped entirely. Finally, the output is normalized. It has been shown that residual connections improve network training and performance, and allow for far deeper networks \cite{he_deep_2015}. The input and output size of all \acp*{GRN} was chosen to be similar to $d_{model}$ (512) and the dense layers in the \ac*{GRN} were of dimension 2048.

After the \ac*{GRN} encoders, the microstructure information was added again to the input path, followed by normalization and positional encoding. The encoded source (strain) and right-shifted target (stress) sequences were then passed through the transformer decoder. The right-shifted stress sequence required as input to the decoder was generated by inserting a zero vector at the first position of the stress sequence and truncating the sequence at the second to last element. After the transformer decoder, two stacked \acp*{GRN} with the microstructure as context were used followed by a linear transformation to generate the stress predictions. 

The model architecture was implemented in \textit{PyTorch} \cite{adam19} using the \textit{PyTorch Lightning} interface. 

Surrogate model parameters are summarized in Tab.~\ref{tab:transformer_architecture}. 
\begin{table}[!htb]
	\centering
	\caption{Transformer architecture parameters.}
	\label{tab:transformer_architecture}
	\begin{tabular}{lc}
		\toprule
		Model parameter & Size \\\midrule
		Transformer size ($d_{model}$) & 512\\
		Transformer feed-forward size & 2048\\
		Number of decoder layers & 6\\
		Number of attention heads & 8\\
		Dropout probability & 0.1\\
		GRN encoder / decoder size & 2048\\
		Number of stacked \acp*{GRN} in encoder / decoder & 2\\
		GRN input / output size & 512\\
		\bottomrule
	\end{tabular}
\end{table}

With \ac*{CNN} microstructure encoding, the chosen network had a total of 66.3\,million trainable parameters. The total trainable parameters consisted of 19\,million parameters for the \ac*{CNN} encoding, 25.2\,million parameters for the transformer decoder, and 7.4\,million parameters for each \ac*{GRN} encoder (including the linear transformation of the strain and stress sequences, respectively) and 7.3\,million parameters for the \ac*{GRN} decoder.

Following \cite{vasvani17}, the transformer model was trained using the \text{Adam} optimizer \cite{kingma15} with learning rate warmup. During training, the learning rate $lr$ was initially increased linearly, followed by decreasing learning rate proportional to the inverse square root of the current step number \cite{vasvani17}. The learning rate was calculated by:

 \begin{equation}
	lr=d_{model}^{-0.5} \min\left( step_{num}^{-0.5}, step_{num}\cdot warmup_{steps}^{-1.5}\right),
\end{equation}

where $d_{model}$ is the attention mechanism dimension, $step_{num}$ is the current training step number and $warmup_{steps}$ is the number of steps of linearly increasing learning rate. As was chosen in \cite{vasvani17}, $warmup_{steps}$ was set to 4000.
Using learning rate warmup can prevent divergent training in transformer models \cite{popel_training_2018}.

Target and source sequences were standardized to zero mean and unit variance before training to speed up the training process. The training sequences were truncated at random indices to allow the transformer to adapt to different length sequences, resulting in sequence lengths between one and 100. As a loss function, the \ac*{MSE} between the predicted and true values was used. It should be noted that due to the choice of loss function, the training process was dominated by high-stress data \cite{wu20}. Therefore, relative prediction performance might decrease for smaller stresses. This effect should be reduced through normalization of the data and the loss function was, therefore, deemed suitable.

\subsection{Training the Surrogate Model on Homogeneous Data}\label{sec:homogeneous_training}

To test the feasibility of a transformer network as a surrogate model for history-dependent plasticity, initially, the model was trained on the response of a purely homogeneous elastoplastic material. Using the strain sequence generation described in Sec.~\ref{sec:data_generation} and a (semi) analytical implementation of incremental J2-plasticity, a large amount of data could be generated at a low computational cost. This way and using the matrix material parameters listed in Tab.~\ref{tab:train_range}, 500,000 stress-strain sequences with a maximum sequence length of 100 were generated. Half the generated sequences consisted of random strain walks while the remaining sequences were cyclic. During training, 80\,\% of the data were used for training and the remaining 20\,\% were used for model validation. The surrogate model architecture remained unchanged except for the \ac*{GRN} context input (microstructure) and the microstructure addition step after the \ac*{GRN} encoders were removed. The training was performed for 800 epochs on four \textit{NVIDIA RTX A6000} GPUs with a batch size of 500. Total training time was approximately 24 hours.

\begin{figure}[!htb]
	\centering
	\begin{subfigure}{.6\linewidth}
		\centering
		\includegraphics{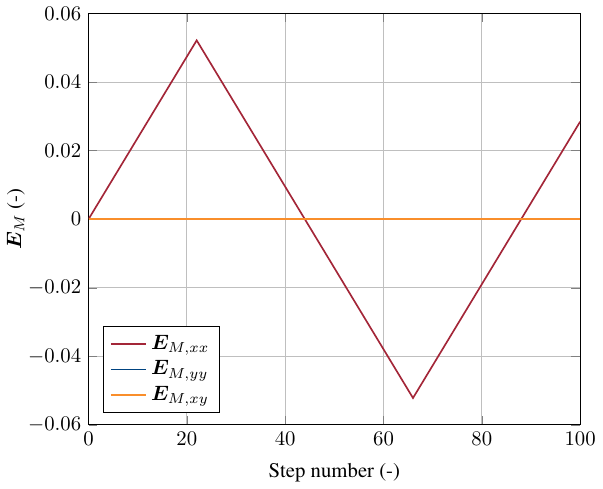}
		\caption{Strain sequence.}
		\label{strain_seq}
	\end{subfigure}
	\begin{subfigure}{.6\linewidth}
		\centering
		\includegraphics{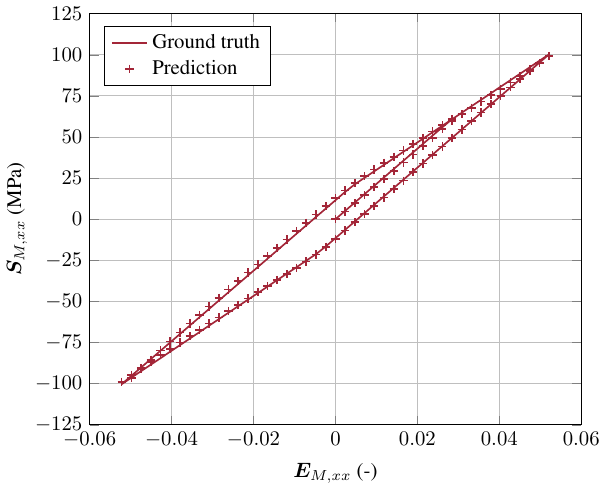}
		\caption{Material response.}
		\label{stress_seq}
	\end{subfigure}
	\caption{Cyclic input strain sequence (\ref{strain_seq}) and comparison between surrogate model prediction and ground truth (\ref{stress_seq}).}
	\label{fig:predictions_homogeneous}
\end{figure}

\begin{figure}[!htb]
	\centering
	\begin{subfigure}{.6\linewidth}
		\centering
		\includegraphics{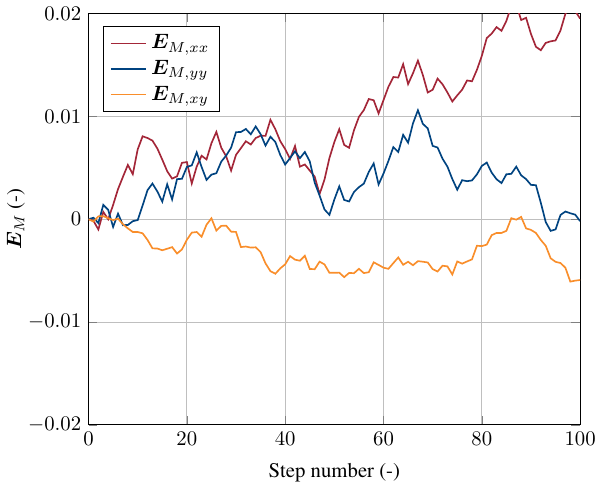}
		\caption{Strain sequence.}
		\label{strain_seq_rand}
	\end{subfigure}
	\begin{subfigure}{.6\linewidth}
		\centering
		\includegraphics{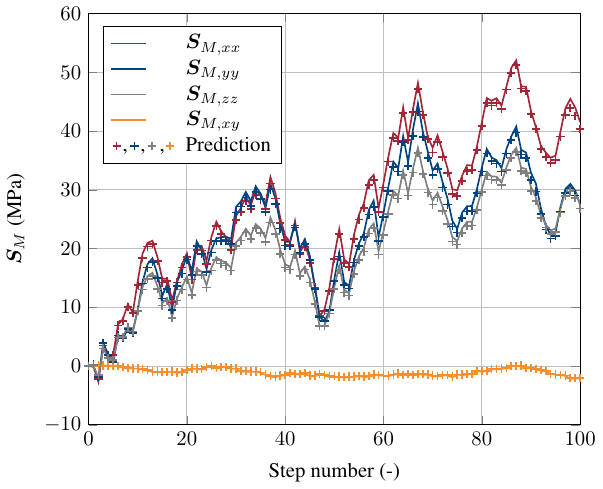}
		\caption{Material response.}
		\label{stress_seq_rand}
	\end{subfigure}
	\caption{Random input strain sequence (\ref{strain_seq_rand}) and comparison between surrogate model prediction and ground truth (\ref{stress_seq_rand}).}
	\label{fig:predictions_homogeneous_rand}
\end{figure}

After 800 epochs, the lowest validation \ac*{MSE} was $5.86\times 10^{-4}$. Note that the \ac*{MSE} is for the standardized dataset and not the actual stress prediction. Subsequently, an additional 500 random and 500 cyclic strain sequences were generated to test the model performance on previously unseen data. Using the trained network the average \ac*{RMSE} of the stress predictions for the ground truth were calculated for the unscaled (non-standardized) data to be 0.804\,MPa. Comparisons between the network predictions and the ground truth data can be seen in Fig.~\ref{fig:predictions_homogeneous} for a cyclic strain sequence and in Fig.~\ref{fig:predictions_homogeneous_rand} for a random strain sequence. History-dependence of the material response for both cyclic (Fig.~\ref{fig:predictions_homogeneous}) and random strain (Fig.~\ref{fig:predictions_homogeneous_rand}) sequences was correctly predicted by the network with only some minor visible differences between the predicted and the true values. 

\subsection{Adapting the Pre-Trained Surrogate Model for Homogenization}

After training the surrogate model for a homogeneous elastoplastic material, we generated a total of 330,000 unique homogenized stress-strain sequences as detailed in Sec.~\ref{sec:data_generation}. This dataset was derived from approximately 33,000 unique two-dimensional \acp*{SVE}, each modeling the transversal behavior of continuously reinforced composites under plane strain conditions. For each unique \ac*{SVE}, around 10 distinct strain sequences were applied, each consisting of 100 strain increments.

The \ac*{FVR} of the generated \acp*{SVE} was varied between 0.2 and 0.5 (the true values vary somewhat, as the target value cannot necessarily be achieved by randomly placing circular fibers inside of a square microcell) with fixed microcell size and fiber diameter. Around 40\,\% of the generated sequences were random walks and the remaining sequences had cyclic loading paths. The material parameters were left constant. An overview of the material parameters and \ac*{SVE} properties used to generate the training data are listed in Tab.~\ref{tab:train_range}. Simulating 330,000 strain sequences took approximately 4 weeks on two \textit{AMD EPYC 7513 32-Core} processors.
\begin{table}[!htb]
	\centering
	\caption{Material and microcell parameters used to generate the training data.}
	\label{tab:train_range}
	\begin{tabular}{lc}
		\toprule
		\ac*{SVE} size ($\mu$m) & 70 \\
		Fiber diameter ($\mu$m) & 7 \\
		\ac*{FVR} (-) & 0.2 -- 0.5\\
		Young's modulus matrix (MPa) & 1000\\
		Poisson's ratio matrix (-) & 0.4\\
		Yield stress matrix (MPa) & 20\\
		Young's modulus fiber (GPa) & 200\\
		Poisson's ratio fiber (-) & 0.18\\
		\bottomrule
	\end{tabular}
\end{table}

Using the generated dataset, two surrogate models with different microstructure representations (Sec.~\ref{sec:microstructure_representation}) were then trained. One of the models used a \ac*{CNN} with 3 output parameters, while the other had as input the first 3 \ac*{PCA} scores of the respective microstructure. Both models were initialized using the pre-trained weights from the homogeneous model and subsequently trained for 50 epochs. Because large batch sizes commonly lead to degradation in generalization performance \cite{hoffer_train_2018}, a smaller batch size of 100 strain sequences was used. Furthermore, the learning rate was reduced by an order of magnitude to avoid ``forgetting'' all previously learned information. Other training parameters remained unchanged from the ones listed in Sec.~\ref{sec:homogeneous_training}.

After 50 training epochs, the lowest \ac*{MSE} on the validation dataset was $2.30\times 10^{-3}$ for the standardized data after epoch 39 when using a \ac*{CNN} for microstructure feature extraction, and $3.04\times 10^{-3}$ after epoch 15 using two-point correlations. Longer required training time to reach minimum \ac*{MSE} when using the \ac*{CNN} can be explained by the randomly initialized \ac*{CNN} weights, while the transformer weights were initialized using the pre-trained weights. Therefore, during training likely mainly the \ac*{CNN} parameters had to be optimized. Larger \ac*{MSE} compared to homogeneous data (minimum homogeneous validation \ac*{MSE} $8.14\times 10^{-5}$) showed the difficulty of the network to adapt to heterogeneous data. This was expected to a certain degree, as the homogenization problem represents inherently complex compound elastoplastic behavior, with the fibers storing significant amounts of elastic energy while plastic deformation occurs in the matrix phase \cite{mozaffar19}.

To evaluate the impact of using pre-trained weights, the surrogate model using \ac*{PCA} was also trained using random weight initialization. The minimum validation loss with random initialization of $3.34\times 10^{-3}$ was reached after 46 epochs, representing considerably longer required training time and a marginally worse minimum loss.

\begin{figure}[!htb]
	\centering
	\includegraphics{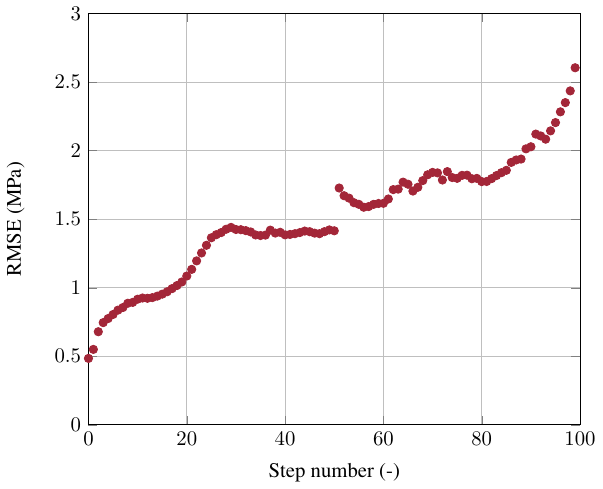}
	\caption{Average \ac*{RMSE} as a function of number of strain increments using two-point correlation microstructure encoding.}
	\label{fig:rmse}
\end{figure}

A test dataset containing 3000 strain sequences for random microstructures was generated to validate the performance of the trained surrogate model for previously unseen data. The \ac*{RMSE} between stress predictions and \ac*{FEM} results on the test data was 2.16\,MPa using the \ac*{CNN}, 1.51\,MPa using \ac*{PCA} with pre-training, and 2.63\,MPa using \ac*{PCA} with random weight initialization. Stresses in the test dataset ranged between -387.5\,MPa and 360.9\,MPa. Both microstructure representation methods were able to successfully relate microstructure information to homogenized properties, with somewhat lower prediction error when using the \ac*{PCA}. Random weight initialization also resulted in worse performance compared to using the pre-trained weights from homogeneous training. The prediction error incrementally rises with increasing sequence length, due to the transformer's approach of predicting one time-step at a time, relying on the preceding data (see Fig.~\ref{fig:rmse}). 

\begin{figure}[!htb]
	\centering
	\begin{subfigure}{.6\linewidth}
		\centering
		\includegraphics{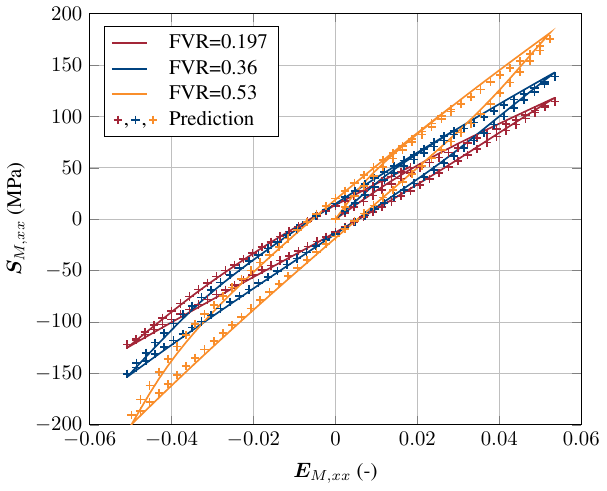}
		\caption{\ac*{CNN} microstructure encoding.}
		\label{fig:cnn_predictions}
	\end{subfigure}
	\begin{subfigure}{.6\linewidth}
		\centering
		\includegraphics{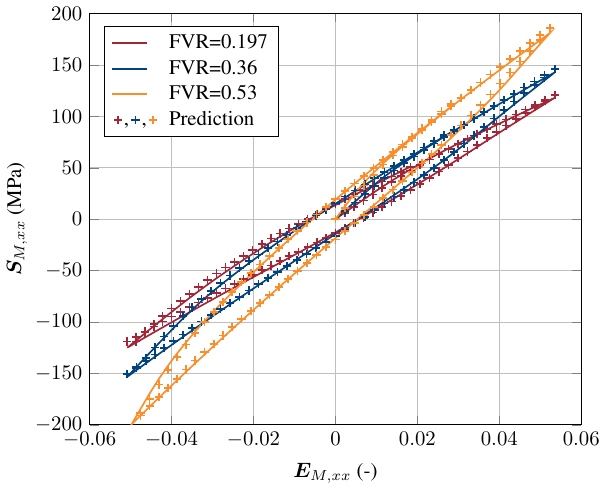}
		\caption{\ac*{PCA} microstructure encoding.}
		\label{fig:pca_predictions}
	\end{subfigure}
	\caption{Comparison between surrogate model prediction and \ac*{FEM} for cyclic loading paths at three different \acp*{FVR} using \ac*{CNN} (\ref{fig:cnn_predictions}) and \ac*{PCA} (\ref{fig:pca_predictions}).}
	\label{fig:predictions_heterogeneous}
\end{figure}

Fig.~\ref{fig:predictions_heterogeneous} demonstrates the comparison between cyclic \ac*{FEM} sequences and the surrogate model performance using both microstructure encoding methods (\ac*{CNN} Fig.~\ref{fig:cnn_predictions}, two-point statistics Fig.~\ref{fig:pca_predictions}). Generally, good accuracy between the \ac*{FEM} and predicted curves can be seen. The surrogate models correctly capture changing stiffness with varying \ac*{FVR} and accurately predict the material history-dependence.

\begin{figure}[!htb]
	\centering
	\begin{subfigure}{.6\linewidth}
		\centering
		\includegraphics{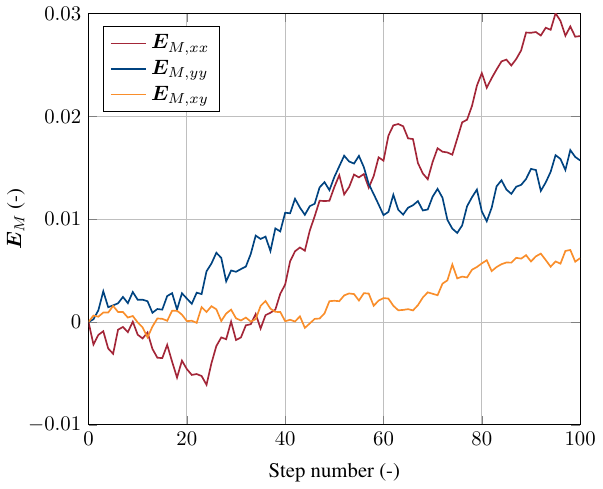}
		\caption{Strain sequence.}
		\label{strain_step_rand}
	\end{subfigure}
	\begin{subfigure}{.6\linewidth}
		\centering
		\includegraphics{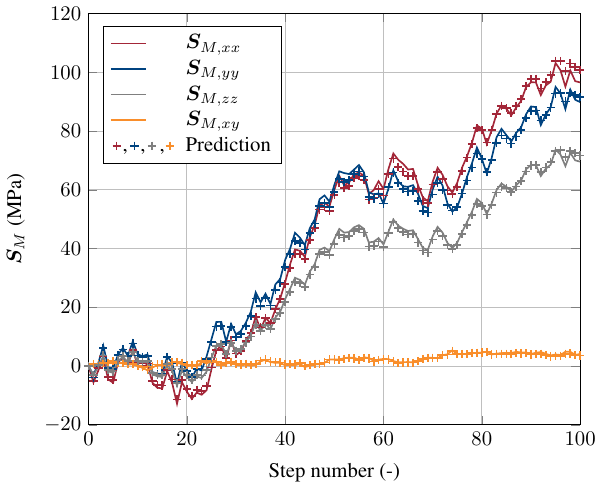}
		\caption{Material response.}
		\label{stress_seq_het_rand}
	\end{subfigure}
	\caption{Random input strain sequence (\ref{strain_step_rand}) and comparison between surrogate model prediction using \ac*{PCA} microstructure representation and ground truth (\ref{stress_seq_het_rand}).}
	\label{fig:predictions_heterogeneous_rand}
\end{figure}

Predictions of the surrogate model using two-point correlations for a random strain sequence are depicted in Fig.~\ref{fig:predictions_heterogeneous_rand}. Generally, a good agreement between surrogate model predictions and \ac*{FEM} results can be seen, including for the out-of-plane stress $\bm{S}_{M,zz}$ component. 

However, several strain sequences from the test dataset, particularly random walks, showed relatively large prediction errors. The fact that the majority of training sequences were cyclic loading paths might be a contributing factor to these outliers. Furthermore, as transformers commonly require large amounts of data \cite{dedhia_scout_2022}, increasing the number of training sequences could improve prediction performance. In this effort, a relatively large model (around 50\,million parameters) was selected. Large models, when trained on small datasets, are more prone to overfitting, leading to poor generalization performance. Optimization of the network size and hyperparameter tuning could further improve the prediction performance of the surrogate models, reduce the training duration, and lower the required training data.

Three principal components and three output parameters of a \ac*{CNN} were used to encode microstructure information as input to the surrogate model. However, it was shown (see Sec.~\ref{sec:microstructure_representation}) that likely a single parameter, namely the encoded \ac*{FVR}, would have been sufficient for the analyzed microstructures. Increasing the complexity of the microstructures used for training by, for example, varying the \ac*{RVE} size and fiber diameter, or introducing non-circular inclusions will require more parameters to encode \ac*{RVE} features. Similarly, higher fidelity damage models to account for matrix damage will also require more refined structure descriptions to account for fiber clusters and voids initiating and accumulating local damage.

Comparing computational cost, solving a single microstructure subject to a random strain sequence with a length of 100 increments using parallel \ac*{FEM} on two CPUs took approximately 40\,s. In contrast, predicting the response of 1000 microcells on a \textit{NVIDIA RTX A6000} GPU using the surrogate model took less than 25\,s. Due to the fast batch processing of the neural network, major time advantages can especially be achieved when predicting a large number of sequences, as would be the case in structural FE$^2$ analysis.

\subsection{Testing Prediction Performance on Unseen Microstructure}

\begin{figure}[!htb]
	\centering
	\includegraphics{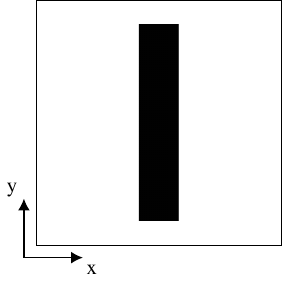}
		\caption{Microcell with rectangular inclusion.}
	\label{fig:square_micro}
\end{figure}

\begin{figure}[!htb]
	\centering
	\begin{subfigure}{.6\linewidth}
		\centering
		\includegraphics{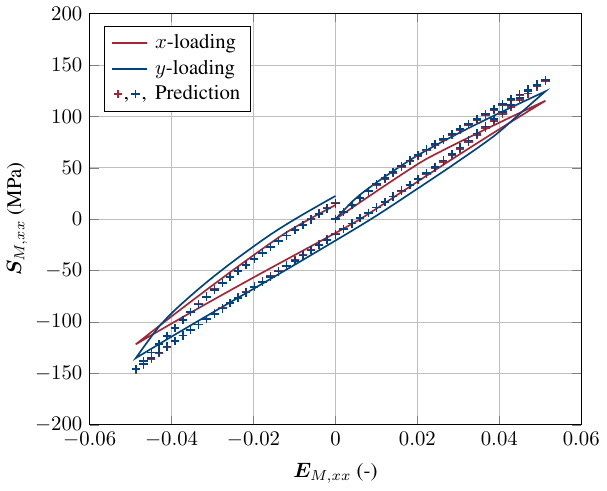}
		\caption{\ac*{CNN} microstructure encoding.}
		\label{fig:square_cnn}
	\end{subfigure}
	\begin{subfigure}{.6\linewidth}
		\centering
		\includegraphics{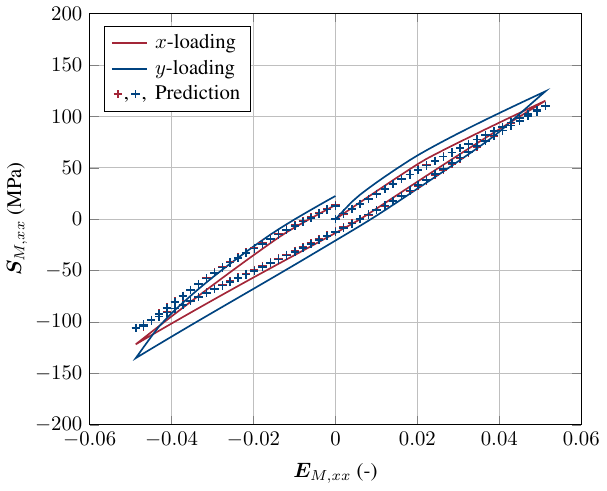}
		\caption{\ac*{PCA} microstructure encoding.}
		\label{fig:square_pca}
	\end{subfigure}
	\caption{Comparison between surrogate model prediction and \ac*{FEM} for cyclic loading paths of an unseen microstructure using \ac*{CNN} (\ref{fig:square_cnn}) and \ac*{PCA} (\ref{fig:square_pca}).}
	\label{fig:square_inclusion}
\end{figure}

To acquire additional insight into the knowledge assimilated by the neural network and test the model's robustness, a completely new microstructure containing a rectangular inclusion (see Fig.~\ref{fig:square_micro}) was simulated under cyclic loading in $x$ and $y$-directions using \ac*{FEM} and the results were compared with the network predictions. The rectangular inclusion was centered inside the microcell with a \ac*{FVR} of 0.2 and a width ($x$) to height ($y$) ratio of $0.25:0.8$. Due to the elongated inclusion shape, anisotropy of the response was introduced (see Fig.~\ref{fig:square_inclusion}). Because the surrogate models were only trained on circular inclusions, the prediction performance for new microstructures can give information about the generalizability of the models.

Comparing the simulated data with the neural network, both microstructure encoding methods were unable to capture the microcell's anisotropic response. Instead, the surrogate models predicted almost identical responses for transversal and longitudinal loading. This was expected as the networks were trained on data that was mostly isotropic and the network predictions were predominantly based on the \ac*{FVR}. However, the \ac*{CNN} seemed to be able to extract information about the \ac*{FVR} even for completely unseen microstructures (see Fig.~\ref{fig:square_cnn}). Using two-point statistics, the transversal response of the microstructure was captured relatively well (see Fig.~\ref{fig:square_pca}) with a larger error for longitudinal loading. This shows that the surrogate models were able to learn a \ac*{FVR}-based homogenization for elastoplastic composites. Additional training on a wider variety of microstructures would allow the models to generalize and give improved predictions for a wide range of inclusion geometries.

\section{Concluding Remarks}
A surrogate model to predict the history-dependent homogenized response of elastoplastic composite \acp*{SVE} based on a \ac*{NLP} transformer neural network has been presented. A decoder-only architecture with \acp*{GRN} to encode time-dependent (strain, stress) and static data (microstructure information) was chosen. Features were generated from images of the material microstructure comparing (i) a learned auto-encoding based on a \ac*{CNN} and (ii) two-point correlation functions using \ac*{PCA} for dimensionality reduction.

After training the developed network using random strain walks and cyclic loading paths, the model successfully predicted the history-dependent response of previously unseen composite \acp*{SVE}. Furthermore, both microstructure representation techniques were able to extract parameters from the microstructure that governed the material response.

Considerable savings in computational cost for microstructure homogenization compared to \ac*{FEM} were demonstrated. Efficiency gains compared to FE$^2$ make \ac*{UQ} of composite structures with randomness in the underlying constituent distribution attainable. The transformer surrogate model could be used for scale transition to estimate the local homogenized material response based on the micro-scale constituent composition and included in a \ac*{MC} or \ac*{QMC} loop to sample the response distribution.

The current implementation was restricted to simple two-dimensional composite \acp*{SVE} of fixed size and fiber diameter and fixed constituent properties. Having shown the feasibility of using a transformer surrogate for homogenization, the model will be extended to account for more complex microstructures, such as randomly sized \acp*{SVE} and non-circular inclusions, and varying material parameters. Extension to three-dimensional microstructures is possible by using a three-dimensional \ac*{CNN} or three-dimensional two-point correlations to encode the microstructure. 

While deep learning has been shown to have the ability to mitigate the curse of dimensionality \cite{poggio17}, multiple strategies will be employed to additionally reduce the required data when training the stress transformer on more complex microstructures: 1) In \cite{bessa_framework_2017}, the authors implement a data-driven material discovery approach using feature importance scores of microstructure parameters to more efficiently cover the training space, thereby escaping the curse of dimensionality. Following their approach, the amount of required training data to extend the transformer model to more complex microstructures could be reduced by carefully designing experiments based on feature importance. 2) Physics-informed loss functions could decrease the required training data amount \cite{nascimento_tutorial_2020}. 3) Pre-training our model on simpler data and then fine-tuning it with complex, higher-dimensional data has proven effective in reducing overall data requirements. 4) Finally, the typically narrow range of modulus and strength ratios between matrix and inclusion in composites limits the parameter space, potentially reducing the volume of data needed for model training.
	
	\section*{Funding Sources}
	This research did not receive any specific grant from funding agencies in the public, commercial, or not-for-profit sectors.
	
	\FloatBarrier
	
	\bibliography{myrefs.bib}

\begin{thebibliography}{10}
\expandafter\ifx\csname url\endcsname\relax
  \def\url#1{\texttt{#1}}\fi
\expandafter\ifx\csname urlprefix\endcsname\relax\def\urlprefix{URL }\fi
\expandafter\ifx\csname href\endcsname\relax
  \def\href#1#2{#2} \def\path#1{#1}\fi

\bibitem{zhang_using_2020}
A.~Zhang, D.~Mohr, Using neural networks to represent von mises plasticity with
  isotropic hardening, International Journal of Plasticity 132 (2020) 102732.
\newblock \href {https://doi.org/10.1016/j.ijplas.2020.102732}
  {\path{doi:10.1016/j.ijplas.2020.102732}}.

\bibitem{bessa_framework_2017}
M.~Bessa, R.~Bostanabad, Z.~Liu, A.~Hu, D.~W. Apley, C.~Brinson, W.~Chen, W.~K.
  Liu, A framework for data-driven analysis of materials under uncertainty:
  Countering the curse of dimensionality, Computer Methods in Applied Mechanics
  and Engineering 320 (2017) 633--667.
\newblock \href {https://doi.org/10.1016/j.cma.2017.03.037}
  {\path{doi:10.1016/j.cma.2017.03.037}}.

\bibitem{furukawa_accurate_2004}
T.~Furukawa, M.~Hoffman, Accurate cyclic plastic analysis using a neural
  network material model, Engineering Analysis with Boundary Elements 28~(3)
  (2004) 195--204.
\newblock \href {https://doi.org/10.1016/S0955-7997(03)00050-X}
  {\path{doi:10.1016/S0955-7997(03)00050-X}}.

\bibitem{chen_polyconvex_2022}
P.~Chen, J.~Guilleminot, Polyconvex neural networks for hyperelastic
  constitutive models: A rectification approach, Mechanics Research
  Communications 125 (2022) 103993.
\newblock \href {https://doi.org/10.1016/j.mechrescom.2022.103993}
  {\path{doi:10.1016/j.mechrescom.2022.103993}}.

\bibitem{wang_meta-modeling_2019}
K.~Wang, W.~Sun, Meta-modeling game for deriving theory-consistent,
  microstructure-based traction–separation laws via deep reinforcement
  learning, Computer Methods in Applied Mechanics and Engineering 346 (2019)
  216--241.
\newblock \href {https://doi.org/10.1016/j.cma.2018.11.026}
  {\path{doi:10.1016/j.cma.2018.11.026}}.

\bibitem{messner_convolutional_2020}
M.~C. Messner, Convolutional neural network surrogate models for the mechanical
  properties of periodic structures, Journal of Mechanical Design 142~(2)
  (2020) 024503.
\newblock \href {https://doi.org/10.1115/1.4045040}
  {\path{doi:10.1115/1.4045040}}.

\bibitem{rao_three-dimensional_2020}
C.~Rao, Y.~Liu, Three-dimensional convolutional neural network (3d-{CNN}) for
  heterogeneous material homogenization, Computational Materials Science 184
  (2020) 109850.
\newblock \href {https://doi.org/10.1016/j.commatsci.2020.109850}
  {\path{doi:10.1016/j.commatsci.2020.109850}}.

\bibitem{peng_ph-net_2022}
H.~Peng, A.~Liu, J.~Huang, L.~Cao, J.~Liu, L.~Lu, {PH}-net: Parallelepiped
  microstructure homogenization via 3d convolutional neural networks, Additive
  Manufacturing 60 (2022) 103237.
\newblock \href {https://doi.org/10.1016/j.addma.2022.103237}
  {\path{doi:10.1016/j.addma.2022.103237}}.

\bibitem{lu_data-driven_2019}
X.~Lu, D.~G. Giovanis, J.~Yvonnet, V.~Papadopoulos, F.~Detrez, J.~Bai, A
  data-driven computational homogenization method based on neural networks for
  the nonlinear anisotropic electrical response of graphene/polymer
  nanocomposites, Computational Mechanics 64~(2) (2019) 307--321.
\newblock \href {https://doi.org/10.1007/s00466-018-1643-0}
  {\path{doi:10.1007/s00466-018-1643-0}}.

\bibitem{le_computational_2015}
B.~A. Le, J.~Yvonnet, Q.~He, Computational homogenization of nonlinear elastic
  materials using neural networks, International Journal for Numerical Methods
  in Engineering 104~(12) (2015) 1061--1084.
\newblock \href {https://doi.org/10.1002/nme.4953}
  {\path{doi:10.1002/nme.4953}}.

\bibitem{nguyen-thanh_surrogate_2019}
V.~M. Nguyen-Thanh, L.~T.~K. Nguyen, T.~Rabczuk, X.~Zhuang, A surrogate model
  for computational homogenization of elastostatics at finite strain using the
  {HDMR}-based neural network approximator (2019).
\newblock \href {http://arxiv.org/abs/1906.02005} {\path{arXiv:1906.02005}}.

\bibitem{mianroodi_teaching_2021}
J.~R. Mianroodi, N.~H.~Siboni, D.~Raabe, Teaching solid mechanics to artificial
  intelligence—a fast solver for heterogeneous materials, npj Computational
  Materials 7~(1) (2021) 99.
\newblock \href {https://doi.org/10.1038/s41524-021-00571-z}
  {\path{doi:10.1038/s41524-021-00571-z}}.

\bibitem{khorrami_artificial_2022}
M.~S. Khorrami, J.~R. Mianroodi, N.~H. Siboni, P.~Goyal, B.~Svendsen,
  P.~Benner, D.~Raabe, An artificial neural network for surrogate modeling of
  stress fields in viscoplastic polycrystalline materials (2022).
\newblock \href {http://arxiv.org/abs/2208.13490} {\path{arXiv:2208.13490}}.

\bibitem{wu20}
L.~Wu, V.~D. Nguyen, N.~G. Kilingar, L.~Noels, A recurrent neural
  network-accelerated multi-scale model for elasto-plastic heterogeneous
  materials subjected to random cyclic and non-proportional loading paths,
  Computer Methods in Applied Mechanics and Engineering 369 (2020) 113234.
\newblock \href {https://doi.org/10.1016/j.cma.2020.113234}
  {\path{doi:10.1016/j.cma.2020.113234}}.

\bibitem{settgast_hybrid_2020}
C.~Settgast, G.~Hütter, M.~Kuna, M.~Abendroth, A hybrid approach to simulate
  the homogenized irreversible elastic–plastic deformations and damage of
  foams by neural networks, International Journal of Plasticity 126 (2020)
  102624.
\newblock \href {https://doi.org/10.1016/j.ijplas.2019.11.003}
  {\path{doi:10.1016/j.ijplas.2019.11.003}}.

\bibitem{mozaffar19}
M.~Mozaffar, R.~Bostanabad, W.~Chen, K.~Ehmann, J.~Cao, M.~A. Bessa, Deep
  learning predicts path-dependent plasticity, Proceedings of the National
  Academy of Sciences 116~(52) (2019) 26414--26420.
\newblock \href {https://doi.org/10.1073/pnas.1911815116}
  {\path{doi:10.1073/pnas.1911815116}}.

\bibitem{ghavamian_accelerating_2019}
F.~Ghavamian, A.~Simone, Accelerating multiscale finite element simulations of
  history-dependent materials using a recurrent neural network, Computer
  Methods in Applied Mechanics and Engineering 357 (2019) 112594.
\newblock \href {https://doi.org/10.1016/j.cma.2019.112594}
  {\path{doi:10.1016/j.cma.2019.112594}}.

\bibitem{wu_recurrent_2022}
L.~Wu, L.~Noels, Recurrent neural networks ({RNNs}) with dimensionality
  reduction and break down in computational mechanics; application to
  multi-scale localization step, Computer Methods in Applied Mechanics and
  Engineering 390 (2022) 114476.
\newblock \href {https://doi.org/10.1016/j.cma.2021.114476}
  {\path{doi:10.1016/j.cma.2021.114476}}.

\bibitem{vasvani17}
A.~Vaswani, N.~Shazeer, N.~Parmar, J.~Uszkoreit, L.~Jones, A.~N. Gomez, L.~u.
  Kaiser, I.~Polosukhin, Attention is all you need, in: I.~Guyon, U.~V.
  Luxburg, S.~Bengio, H.~Wallach, R.~Fergus, S.~Vishwanathan, R.~Garnett
  (Eds.), Advances in Neural Information Processing Systems, Vol.~30, 2017.
\newblock \href {http://arxiv.org/abs/1706.03762} {\path{arXiv:1706.03762}}.

\bibitem{radford_language_2019}
A.~Radford, J.~Wu, R.~Child, D.~Luan, D.~Amodei, I.~Sutskever, Language models
  are unsupervised multitask learners (2019).

\bibitem{devlin_bert_2019}
J.~Devlin, M.-W. Chang, K.~Lee, K.~Toutanova, {BERT}: Pre-training of deep
  bidirectional transformers for language understanding (2019).
\newblock \href {http://arxiv.org/abs/1810.04805} {\path{arXiv:1810.04805}}.

\bibitem{radford_improving_2018}
A.~Radford, K.~Narasimhan, Improving language understanding by generative
  pre-training, 2018.

\bibitem{floridi_gpt-3_2020}
L.~Floridi, M.~Chiriatti, {GPT}-3: Its nature, scope, limits, and consequences,
  Minds and Machines 30~(4) (2020) 681--694.
\newblock \href {https://doi.org/10.1007/s11023-020-09548-1}
  {\path{doi:10.1007/s11023-020-09548-1}}.

\bibitem{openai23}
J.~Achiam, S.~Adler, S.~Agarwal, L.~Ahmad, I.~Akkaya, F.~L. Aleman, et~al.,
  {GPT}-4 {T}echnical {R}eport (2023).
\newblock \href {http://arxiv.org/abs/2303.08774} {\path{arXiv:2303.08774}}.

\bibitem{lim_temporal_2021}
B.~Lim, S.~O. Arik, N.~Loeff, T.~Pfister, Temporal fusion transformers for
  interpretable multi-horizon time series forecasting, International Journal of
  Forecasting 37~(4) (2021) 1748--1764.
\newblock \href {https://doi.org/10.1016/j.ijforecast.2021.03.012}
  {\path{doi:10.1016/j.ijforecast.2021.03.012}}.

\bibitem{eisenstein_artificial_2021}
M.~Eisenstein, Artificial intelligence powers protein-folding predictions,
  Nature 599~(7886) (2021) 706--708.
\newblock \href {https://doi.org/10.1038/d41586-021-03499-y}
  {\path{doi:10.1038/d41586-021-03499-y}}.

\bibitem{alquraishi_machine_2021}
M.~{AlQuraishi}, Machine learning in protein structure prediction, Current
  Opinion in Chemical Biology 65 (2021) 1--8.
\newblock \href {https://doi.org/10.1016/j.cbpa.2021.04.005}
  {\path{doi:10.1016/j.cbpa.2021.04.005}}.

\bibitem{jumper_highly_2021}
J.~Jumper, R.~Evans, A.~Pritzel, T.~Green, M.~Figurnov, O.~Ronneberger,
  K.~Tunyasuvunakool, R.~Bates, A.~Žídek, A.~Potapenko, A.~Bridgland,
  C.~Meyer, S.~A.~A. Kohl, A.~J. Ballard, A.~Cowie, B.~Romera-Paredes,
  S.~Nikolov, R.~Jain, J.~Adler, T.~Back, S.~Petersen, D.~Reiman, E.~Clancy,
  M.~Zielinski, M.~Steinegger, M.~Pacholska, T.~Berghammer, S.~Bodenstein,
  D.~Silver, O.~Vinyals, A.~W. Senior, K.~Kavukcuoglu, P.~Kohli, D.~Hassabis,
  Highly accurate protein structure prediction with {AlphaFold}, Nature
  596~(7873) (2021) 583--589.
\newblock \href {https://doi.org/10.1038/s41586-021-03819-2}
  {\path{doi:10.1038/s41586-021-03819-2}}.

\bibitem{chen_generative_2020}
M.~Chen, A.~Radford, R.~Child, J.~Wu, H.~Jun, D.~Luan, I.~Sutskever, Generative
  pretraining from pixels, in: H.~D. III, A.~Singh (Eds.), Proceedings of the
  37th International Conference on Machine Learning, Vol. 119 of Proceedings of
  Machine Learning Research, PMLR, 2020, pp. 1691--1703.

\bibitem{dosovitskiy_image_2020}
A.~Dosovitskiy, L.~Beyer, A.~Kolesnikov, D.~Weissenborn, X.~Zhai,
  T.~Unterthiner, M.~Dehghani, M.~Minderer, G.~Heigold, S.~Gelly, J.~Uszkoreit,
  N.~Houlsby, An image is worth 16x16 words: Transformers for image recognition
  at scale (2020).
\newblock \href {http://arxiv.org/abs/12010.11929} {\path{arXiv:12010.11929}}.

\bibitem{han_survey_2023}
K.~Han, Y.~Wang, H.~Chen, X.~Chen, J.~Guo, Z.~Liu, Y.~Tang, A.~Xiao, C.~Xu,
  Y.~Xu, Z.~Yang, Y.~Zhang, D.~Tao, A survey on vision transformer, {IEEE}
  Transactions on Pattern Analysis and Machine Intelligence 45~(1) (2023)
  87--110.
\newblock \href {http://arxiv.org/abs/2012.12556} {\path{arXiv:2012.12556}},
  \href {https://doi.org/10.1109/TPAMI.2022.3152247}
  {\path{doi:10.1109/TPAMI.2022.3152247}}.

\bibitem{chen_mesh_transformer_2022}
Y.~Chen, J.~Zhao, L.~Huang, H.~Chen, 3d mesh transformer: A hierarchical neural
  network with local shape tokens, Neurocomputing 514 (2022) 328--340.
\newblock \href {https://doi.org/10.1016/j.neucom.2022.09.138}
  {\path{doi:10.1016/j.neucom.2022.09.138}}.

\bibitem{dedhia_scout_2022}
B.~Dedhia, R.~Balasubramanian, N.~K. Jha, {SCouT}: Synthetic counterfactuals
  via spatiotemporal transformers for actionable healthcare (2022).
\newblock \href {http://arxiv.org/abs/2207.04208} {\path{arXiv:2207.04208}}.

\bibitem{melnychuk_causal_2022}
V.~Melnychuk, D.~Frauen, S.~Feuerriegel, Causal transformer for estimating
  counterfactual outcomes (2022).
\newblock \href {http://arxiv.org/abs/2204.07258} {\path{arXiv:2204.07258}}.

\bibitem{wu_deep_2020}
N.~Wu, B.~Green, X.~Ben, S.~O'Banion, Deep transformer models for time series
  forecasting: The influenza prevalence case (2020).
\newblock \href {http://arxiv.org/abs/2001.08317} {\path{arXiv:2001.08317}}.

\bibitem{li_enhancing_2020}
S.~Li, X.~Jin, Y.~Xuan, X.~Zhou, W.~Chen, Y.-X. Wang, X.~Yan, Enhancing the
  locality and breaking the memory bottleneck of transformer on time series
  forecasting (2020).
\newblock \href {http://arxiv.org/abs/1907.00235} {\path{arXiv:1907.00235}}.

\bibitem{zhou_informer_2021}
H.~Zhou, S.~Zhang, J.~Peng, S.~Zhang, J.~Li, H.~Xiong, W.~Zhang, Informer:
  Beyond efficient transformer for long sequence time-series forecasting,
  Proceedings of the {AAAI} Conference on Artificial Intelligence 35~(12)
  (2021) 11106--11115.
\newblock \href {https://doi.org/10.1609/aaai.v35i12.17325}
  {\path{doi:10.1609/aaai.v35i12.17325}}.

\bibitem{aggarwal_neural_2018}
C.~C. Aggarwal, Neural Networks and Deep Learning: A Textbook, Springer
  International Publishing, 2018.
\newblock \href {https://doi.org/10.1007/978-3-319-94463-0}
  {\path{doi:10.1007/978-3-319-94463-0}}.

\bibitem{bahdanau_neural_2016}
D.~Bahdanau, K.~Cho, Y.~Bengio, Neural machine translation by jointly learning
  to align and translate (2016).
\newblock \href {http://arxiv.org/abs/1409.0473} {\path{arXiv:1409.0473}}.

\bibitem{jain_attention_2019}
S.~Jain, B.~C. Wallace, Attention is not explanation (2019).
\newblock \href {http://arxiv.org/abs/1902.10186} {\path{arXiv:1902.10186}}.

\bibitem{wiegreffe_attention_2019}
S.~Wiegreffe, Y.~Pinter, Attention is not not explanation (2019).
\newblock \href {http://arxiv.org/abs/1908.04626} {\path{arXiv:1908.04626}}.

\bibitem{he_deep_2015}
K.~He, X.~Zhang, S.~Ren, J.~Sun, Deep residual learning for image recognition
  (2015).
\newblock \href {http://arxiv.org/abs/1512.03385} {\path{arXiv:1512.03385}}.

\bibitem{jurafsky_speech_2009}
D.~Jurafsky, J.~H. Martin,
  \href{https://web.stanford.edu/~jurafsky/slp3/}{Speech and language
  processing: an introduction to natural language processing, computational
  linguistics, and speech recognition}, 3rd Edition, 2023.
\newline\urlprefix\url{https://web.stanford.edu/~jurafsky/slp3/}

\bibitem{peric_micro-macro_2011}
D.~Peri\'{c}, E.~A. de~Souza~Neto, R.~A. Feij\'{o}o, M.~Partovi, A.~J.~C.
  Molina, On micro-to-macro transitions for multi-scale analysis of non-linear
  heterogeneous materials: unified variational basis and finite element
  implementation, International Journal for Numerical Methods in Engineering
  87~(1) (2011) 149--170.
\newblock \href {https://doi.org/10.1002/nme.3014}
  {\path{doi:10.1002/nme.3014}}.

\bibitem{wu19}
L.~Wu, V.-D. Nguyen, L.~Adam, L.~Noels, An inverse micro-mechanical analysis
  toward the stochastic homogenization of nonlinear random composites, Computer
  Methods in Applied Mechanics and Engineering 348 (2019) 97--138.
\newblock \href {https://doi.org/10.1016/j.cma.2019.01.016}
  {\path{doi:10.1016/j.cma.2019.01.016}}.

\bibitem{geuzaine_gmsh_2009}
C.~Geuzaine, J.-F. Remacle, Gmsh: A 3-d finite element mesh generator with
  built-in pre- and post-processing facilities, International Journal for
  Numerical Methods in Engineering 79~(11) (2009) 1309--1331.
\newblock \href {https://doi.org/10.1002/nme.2579}
  {\path{doi:10.1002/nme.2579}}.

\bibitem{dhont_calculix}
G.~Dhondt, K.~Wittig, \href{http://www.dhondt.de}{Calculi{X}} (2022).
\newline\urlprefix\url{http://www.dhondt.de}

\bibitem{dhondt_finite_2004}
G.~Dhondt, The Finite Element Method for Three-Dimensional Thermomechanical
  Applications, 1st Edition, Wiley, 2004.
\newblock \href {https://doi.org/10.1002/0470021217}
  {\path{doi:10.1002/0470021217}}.

\bibitem{xu_machine_2015}
H.~Xu, R.~Liu, A.~Choudhary, W.~Chen, A machine learning-based design
  representation method for designing heterogeneous microstructures, Journal of
  Mechanical Design 137~(5) (2015) 051403.
\newblock \href {https://doi.org/10.1115/1.4029768}
  {\path{doi:10.1115/1.4029768}}.

\bibitem{cecen_versatile_2016}
A.~Cecen, T.~Fast, S.~R. Kalidindi, Versatile algorithms for the computation of
  2-point spatial correlations in quantifying material structure, Integrating
  Materials and Manufacturing Innovation 5~(1) (2016) 1--15.
\newblock \href {https://doi.org/10.1186/s40192-015-0044-x}
  {\path{doi:10.1186/s40192-015-0044-x}}.

\bibitem{brough_materials_2017}
D.~B. Brough, D.~Wheeler, S.~R. Kalidindi, Materials knowledge systems in
  python—a data science framework for accelerated development of hierarchical
  materials, Integrating Materials and Manufacturing Innovation 6~(1) (2017)
  36--53.
\newblock \href {https://doi.org/10.1007/s40192-017-0089-0}
  {\path{doi:10.1007/s40192-017-0089-0}}.

\bibitem{ford_machine_2021}
E.~Ford, K.~Maneparambil, S.~Rajan, N.~Neithalath, Machine learning-based
  accelerated property prediction of two-phase materials using microstructural
  descriptors and finite element analysis, Computational Materials Science 191
  (2021) 110328.
\newblock \href {https://doi.org/10.1016/j.commatsci.2021.110328}
  {\path{doi:10.1016/j.commatsci.2021.110328}}.

\bibitem{brough_pymks_online}
D.~B. Brough, D.~Wheeler, S.~R. Kalidindi, \href{pymks.readthedocs.io/}{Pymks
  overview} (2023).
\newline\urlprefix\url{pymks.readthedocs.io/}

\bibitem{hotelling_analysis_1933}
H.~Hotelling, Analysis of a complex of statistical variables into principal
  components., Journal of Educational Psychology 24~(6) (1933) 417--441.
\newblock \href {https://doi.org/10.1037/h0071325}
  {\path{doi:10.1037/h0071325}}.

\bibitem{tipping_mixtures_1999}
M.~E. Tipping, C.~M. Bishop, Mixtures of probabilistic principal component
  analyzers, Neural Computation 11~(2) (1999) 443--482.
\newblock \href {https://doi.org/10.1162/089976699300016728}
  {\path{doi:10.1162/089976699300016728}}.

\bibitem{barnett_prediction_2021}
P.~R. Barnett, S.~A. Young, N.~J. Patel, D.~Penumadu, Prediction of strength
  and modulus of discontinuous carbon fiber composites considering stochastic
  microstructure, Composites Science and Technology 211 (2021) 108857.
\newblock \href {https://doi.org/10.1016/j.compscitech.2021.108857}
  {\path{doi:10.1016/j.compscitech.2021.108857}}.

\bibitem{yang13}
L.~Yang, Y.~Yan, J.~Ma, B.~Liu, Effects of inter-fiber spacing and thermal
  residual stress on transverse failure of fiber-reinforced polymer–matrix
  composites, Computational Materials Science 68 (2013) 255--262.
\newblock \href {https://doi.org/10.1016/j.commatsci.2012.09.027}
  {\path{doi:10.1016/j.commatsci.2012.09.027}}.

\bibitem{krizhevsky17}
A.~Krizhevsky, I.~Sutskever, G.~E. Hinton, Imagenet classification with deep
  convolutional neural networks, Commun. ACM 60~(6) (2017) 84–90.
\newblock \href {https://doi.org/10.1145/3065386} {\path{doi:10.1145/3065386}}.

\bibitem{clevert_fast_2016}
D.-A. Clevert, T.~Unterthiner, S.~Hochreiter, Fast and accurate deep network
  learning by exponential linear units ({ELUs}) (2016).
\newblock \href {http://arxiv.org/abs/1511.07289} {\path{arXiv:1511.07289}}.

\bibitem{dauphin_language_17}
Y.~N. Dauphin, A.~Fan, M.~Auli, D.~Grangier, Language modeling with gated
  convolutional networks, in: D.~Precup, Y.~W. Teh (Eds.), Proceedings of the
  34th International Conference on Machine Learning, Vol.~70 of Proceedings of
  Machine Learning Research, PMLR, 2017, pp. 933--941.

\bibitem{adam19}
A.~Paszke, S.~Gross, F.~Massa, A.~Lerer, J.~Bradbury, G.~Chanan, T.~Killeen,
  Z.~Lin, N.~Gimelshein, L.~Antiga, A.~Desmaison, A.~Kopf, E.~Yang, Z.~DeVito,
  M.~Raison, A.~Tejani, S.~Chilamkurthy, B.~Steiner, L.~Fang, J.~Bai,
  S.~Chintala, Pytorch: An imperative style, high-performance deep learning
  library, in: H.~Wallach, H.~Larochelle, A.~Beygelzimer, F.~d\textquotesingle
  Alch\'{e}-Buc, E.~Fox, R.~Garnett (Eds.), Advances in Neural Information
  Processing Systems 32, Curran Associates, Inc., 2019, pp. 8024--8035.

\bibitem{kingma15}
D.~P. Kingma, J.~Ba, Adam: {A} method for stochastic optimization, in:
  Y.~Bengio, Y.~LeCun (Eds.), 3rd International Conference on Learning
  Representations, {ICLR} 2015, San Diego, CA, USA, May 7-9, 2015, Conference
  Track Proceedings, 2015.

\bibitem{popel_training_2018}
M.~Popel, O.~Bojar, Training tips for the transformer model, The Prague
  Bulletin of Mathematical Linguistics 110~(1) (2018) 43--70.
\newblock \href {http://arxiv.org/abs/1804.00247} {\path{arXiv:1804.00247}},
  \href {https://doi.org/10.2478/pralin-2018-0002}
  {\path{doi:10.2478/pralin-2018-0002}}.

\bibitem{hoffer_train_2018}
E.~Hoffer, I.~Hubara, D.~Soudry, Train longer, generalize better: closing the
  generalization gap in large batch training of neural networks (2018).
\newblock \href {http://arxiv.org/abs/1705.08741} {\path{arXiv:1705.08741}}.

\bibitem{poggio17}
T.~Poggio, H.~Mhaskar, L.~Rosasco, B.~Miranda, Q.~Liao, Why and when can
  deep-but not shallow-networks avoid the curse of dimensionality: A review
  14~(5)  503--519.
\newblock \href {https://doi.org/10.1007/s11633-017-1054-2}
  {\path{doi:10.1007/s11633-017-1054-2}}.

\bibitem{nascimento_tutorial_2020}
R.~G. Nascimento, K.~Fricke, F.~A. Viana, A tutorial on solving ordinary
  differential equations using python and hybrid physics-informed neural
  network, Engineering Applications of Artificial Intelligence 96 (2020)
  103996.
\newblock \href {https://doi.org/10.1016/j.engappai.2020.103996}
  {\path{doi:10.1016/j.engappai.2020.103996}}.

\end{thebibliography}

\end{document}